\documentclass[10pt,journal]{IEEEtran}

\ifCLASSOPTIONcompsoc
  \usepackage[nocompress]{cite}
\else
  \usepackage{cite}
\fi

\ifCLASSINFOpdf
\else
\fi

\usepackage{tikz}
\usetikzlibrary{trees}
\usepackage{pgfplots}
\usepackage[nolist]{acronym} 

\usepackage{mathtools} 
\usepackage{amsmath}
\usepackage{amssymb}
\usepackage{verbatim}
\usepackage{lineno}
\usepackage{latexsym}
\usepackage{times}
\usepackage{bm}

\tikzstyle{every node}=[draw=black,thick,anchor=west]
\tikzstyle{selected}=[draw=red,fill=red!30]
\tikzstyle{optional}=[dashed,fill=gray!50]
\pdfmapfile{=rfxlr-alt.map}
\definecolor{mycolor1}{rgb}{0.00000,0.44700,0.74100}
\definecolor{mycolor2}{rgb}{0.92900,0.69400,0.12500}
\usepackage{hyperref}
\usepackage[utf8]{inputenc}

\usepackage[acronym]{glossaries}
\makeglossaries

 \newacronym{CT}{CT}{Computed Tomography}
\newacronym{PACS}{PACS}{picture archiving and communication system}
\newacronym{MRI}{MRI}{Magnetic Resonance Imaging}
\newacronym{RNA}{RNA}{Ribonucleic Acid}
\newacronym{DUQ}{DUQ}{Deterministic Uncertainty Quantification}
\newacronym{WHO}{WHO}{World Health Organization}
\newacronym{MSE}{MSE}{Mean Squared Error}
\newacronym{TIFF}{TIFF}{tagged image file format}
\newacronym{RT-PCR}{RT-PCR}{ Real-time Reverse Transcription Polymerase Chain Reaction}
\newacronym{COVs}{COVs}{Coronaviruses}
\newacronym{SARS}{SARS}{Severe Acute Respiratory Syndrome}
\newacronym{ARDS}{ARDS}{Acute Respiratory Distress Syndrome}
\newacronym{MERS}{MERS}{Middle East Respiratory Syndrome}

\newacronym{AI}{AI}{Artificial Intelligence}
\newacronym{CNN}{CNN}{Convolutional Neural Network}
\newacronym{CAD}{CAD}{Computer Aided Diagnosis}
\newacronym{OOD}{OOD}{Out of Distribution}
\newacronym{DNN}{DNN}{Deep Neural Network}
\newacronym{IOD}{IOD}{In-Distribution}
\newacronym{GAN}{GAN}{Generative Adversarial Network}
\newacronym{ITU}{ITU}{International Telecommunications Union}

\newacronym{DeDiM}{DeDiM}{Deep Dataset Dissimilarity Measure}
\newacronym{PBC}{PBC}{Pseudo-label based Balance Correction}
\newacronym{SSDL}{SSDL}{Semi-supervised Deep Learning}
\newacronym{UDA}{UDA}{Unsupervised Domain Adaptation}
\newacronym{FH}{FH}{Feature Histograms}

\newacronym{IID}{IID}{Independent and Identically Distributed}
\newacronym{FCN}{FCN}{Fully Connected Network}
\newacronym{MNIST}{MNIST}{Modified National Institute of Standards and Technology dataset}
\newacronym{SVHN}{SVHN}{Street View House Numbers dataset}
\newacronym{CIFAR-10}{CIFAR-10}{Canadian Institute for Advanced Research dataset of 10 classes}

\newacronym{DBN}{DBN}{Deep Belief Network}

\newacronym{TC}{TC}{true certainty}
\newacronym{FC}{FC}{false certainty}
\newacronym{TU}{TU}{true uncertainty}
\newacronym{FU}{FU}{false uncertainty}
\newacronym{TCIA}{TCIA}{The Cancer Imaging Archive}
\newacronym{TP}{TP}{true positives}
\newacronym{ODIN}{ODIN}{Out of DIstribution detector for Neural networks}
\newacronym{TN}{TN}{true negatives}
\newacronym{FP}{FP}{false positives}
\newacronym{FN}{FN}{false negatives}
\newacronym{AUROC}{AUROC}{Area Under the Receiver Operating Characteristic Curve}
\newacronym{DeDiMs}{DeDiMs}{Deep data set Dissimilarity Measures}
\newacronym{SIMD}{SIMD}{Single Input-Multiple Data}
\newacronym{ReLU}{ReLU}{Rectified
Linear Unit}
\newacronym{FE}{FE}{Feature Extractor}
\newacronym{TM}{TM}{Top Model}
\newacronym{PDF}{PDF}{Probability Density Function}
\newacronym{ANOVA}{ANOVA}{Analysis of Variance}
\newacronym{BIRADS}{BIRADS}{Breast Imaging Report and Data System}
\newacronym{DDSM}{DDSM}{Digital
Database for Screening Mammography database}
\newacronym{PA}{PA}{Posteroanterior}
\newacronym{AP}{AP}{Anteroposterior}
\newacronym{LA}{LA}{Lateral}
\newacronym{PCR}{PCR}{Transcription-Polymerase Chain Reaction}
\newacronym{Pi-M}{Pi-M}{Pi Model} 
\newacronym{LaM}{LaM}{Ladder auto-encoder Model} 
\newacronym{TEM}{TEM}{Temporal Ensemble Model} 
\newacronym{VATM}{VATM}{Virtual Adversarial Training Model}
\newacronym{MeM}{MeM}{Memory based Model}
\newacronym{TransM}{TransM}{Transductive Model}
\newacronym{SESEMI}{SESEMI}{Self Supervised network Model}
\newacronym{WaM}{WaM}{Walker Model}
\newacronym{MTM}{MTM}{Mean Teacher Model}
\newacronym{CAT-GAN}{CAT-GAN}{Categorical Generative Adversarial Network}
\newacronym{FM-GAN}{FM-GAN}{Feature Matching Generative Adversarial Network}
\newacronym{Triple-GAN}{Triple-GAN}{Triple Generative Adversarial Network}
\newacronym{Bad-GAN}{Bad-GAN}{Bad Generative Adversarial Network}
\newacronym{Co-GAN}{Co-GAN}{Co-trained Generative Adversarial Network}
\newacronym{Embed-M}{Embed-M}{Embedded Model}
\newacronym{SNTGM}{SNTGM}{Smooth Neighbors on Teacher Graphs Model}
\newacronym{MaxM-GAN}{MaxM-GAN}{Max-min Generative Adversarial Network}
\newacronym{DCNN}{DCNN}{Deep Convolutional Neural Network}
\newacronym{Tri-Net}{Tri-Net}{Tri-net semi-supervised deep model}
\newacronym{DGLN}{DGLN}{Deep Growing Learning Network}
\newacronym{SaaSM}{SaaSM}{Speed as a supervisor for semi-supervised Learning Model}
\newacronym{PT-SSDL}{PT-SSDL}{Pre-trained Semi-Supervised deep learning}
\newacronym{PLT-SSDL}{PLT-SSDL}{Pseudo-label Semi-Supervised deep learning}
\newacronym{R-SSDL}{R-SSDL}{Regularized Semi-Supervised deep learning}
\newacronym{CR-SSDL}{CR-SSDL}{Consistency based Regularized Semi-Supervised deep learning}
\newacronym{GR-SSDL}{GR-SSDL}{Graph based Regularized Semi-Supervised Deep Learning}
\newacronym{ME}{ME}{Mutual Exclusivity}
\newacronym{CC}{CC}{Craniocaudal}
\newacronym{MLO}{MLO}{Mediolateral Oblique}
\newacronym{ACR}{ACR}{American College of Radiology}
\newacronym{CBIS-DDSM}{CBIS-DDSM}{Curated Breast Imaging Subset of Digital Database for Screening Mammography}
\newacronym{RSNA}{RSNA}{Radiological Society of North America}
\newacronym{METM}{METM}{Mutual Exclusivity-Transformation Model}
\newacronym{VATM+EM}{VATM+EM}{Virtual Adversarial Training with Entropy Minimization}
\newacronym{TransM+MTM}{TransM+MTM}{Transductive Model with Mean Teacher}
\newacronym{MEM}{MEM}{Mutual Exclusivity Model}
\newacronym{CTA}{CTA}{CTAugmentation}
\newacronym{RA}{RA}{Random Augmentation}
\newacronym{Co-8-GAN}{Co-8-GAN}{Co-trained Generative Adversarial Network with 8 views}

\newacronym{RC-SSDL}{RC-SSDL}{Consistency Regularized Semi-supervised deep learning}
\newacronym{GaNC-SSDL}{GaNC-SSDL}{Generative adversarial Network based Consistency Regularized Semi-supervised deep learning}
\newacronym{SNTGM+VATM}{SNTGM+VATM}{Smooth Neighbors on Teacher Graphs Model with Virtual Adversarial Training}
\newacronym{SNTGM+Pi-M}{SNTGM+Pi-M}{Smooth Neighbors on Teacher Graphs Model with Pi-Model regularization}
\newacronym{CIFAR-100}{CIFAR-100}{Canadian Institute For Advanced Research dataset with 100 classes}

\newacronym{ILSVRC}{ILSVRC}{ImageNet Large Scale Visual Recognition Challenge}
\newacronym{STL-10}{STL-10}{Self-Taught Learning 10 classes}

\newacronym{TriNet+Pi}{TriNet+Pi}{Tri-net semi-supervised deep model with a Pi-Model regularization}
\newacronym{SVM}{SVM}{Support Vector Machine}
\newacronym{D3SL}{D3SL}{Deep Safe Semi-Supervised Learning}
\newacronym{MTCF}{MTCF}{Multi-Task Curriculum Framework}
\newacronym{TIN}{TIN}{Tiny ImageNet}
\newacronym{LSUN}{LSUN}{Large-scale Scene Understanding dataset }
\newacronym{IF}{IF}{Implicit Differentiation }
\newacronym{MetaA}{MetaA}{Meta Approximation}
\newacronym{UASD}{UASD}{Uncertainty Aware Self-Distillation}
\newacronym{ADA}{ADA}{Augmented Distribution Alignment}
\newacronym{Open-Set-SSLS}{Open-Set-SSLS}{Open-set semi-supervised learning setting}
\newacronym{MCD}{MCD}{Monte Carlo Dropout}
\newacronym{ROC}{ROC}{Receiver Operator Characteristic}
\newacronym{ECE}{ECE}{Expected Calibration Error}
 
\newacronym{RCNN}{RCNN}{Recurrent Convolutional Neural Network}
\newacronym{BIMCV}{BIMCV}{Valencian Region Medical ImageBank}
\newacronym{GN}{GN}{Gaussian Noise}
\newacronym{SP}{SP}{Salt and Pepper}
\newacronym{OH}{OH}{Other-Half}
\newacronym{Dif}{Dif}{Different}
\newacronym{Sim}{Sim}{Similar}
\newacronym{TI}{TI}{Tiny ImageNet}
\newacronym{SAPN}{SAPN}{Salt and Pepper Noise}

\newacronym{TB-1}{TB-1}{Test-bed 1}
\newacronym{TB-1.1}{TB-1.1}{Test-bed 1.1}
\newacronym{TB-1.2}{TB-1.2}{Test-bed 1.2}
\newacronym{TB-2}{TB-2}{Test-bed 2}
\newacronym{JS}{JS}{Jensen-Shannon}

\usepackage{cleveref} 

\begin{document}
 
\title{Semi-supervised Deep Learning for Image Classification with Distribution Mismatch: A Survey}

\author{Saul Calderon-Ramirez, Shengxiang Yang,~\IEEEmembership{Senior Member,~IEEE,} and David Elizondo

\thanks{Manuscript received 07 February 2022.}

\thanks{S. Calderon-Ramirez, S. Yang, and  D. Elizondo are with the Institute of Artificial Intelligence (IAI), De Montfort University, Leicester LE1 9BH,  United Kingdom (e-mail: sacalderon@itcr.ac.cr,  syang@dmu.ac.uk, elizondo@dmu.ac.uk).}}

%
%

\markboth{~Vol.~XX, No.~YY, Month~2022}
{Calderon-Ramirez \MakeLowercase{\textit{et al.}}: Semi-supervised Deep Learning for Image Classification with Distribution Mismatch: A Survey}
%



\IEEEtitleabstractindextext{%
\begin{abstract}
 Deep learning methodologies have been employed in several different fields, with an outstanding success in image recognition applications, such as material quality control, medical imaging, autonomous driving, etc. Deep learning models rely on the abundance of labelled observations to train a prospective model. These models are composed of millions of parameters to estimate, increasing the need of more training observations. Frequently it is expensive to gather labelled observations of data, making the usage of deep learning models not ideal, as the model might over-fit data. In a semi-supervised setting, unlabelled data is used to improve the levels of accuracy and generalization of a model with small labelled datasets. Nevertheless, in many situations different unlabelled data sources might be available. This raises the risk of a significant distribution mismatch between the labelled and unlabelled datasets. Such phenomena can cause a considerable performance hit to typical semi-supervised deep learning frameworks, which often assume that both labelled and unlabelled datasets are drawn from similar distributions. Therefore, in this paper we study the latest approaches for semi-supervised deep learning for image recognition. Emphasis is made in semi-supervised deep learning models designed to deal with a distribution mismatch between the labelled and unlabelled datasets. We address open challenges with the aim to encourage the community to tackle them, and overcome the high data demand of traditional deep learning pipelines under real-world usage settings.

\textit{Impact statement}:  This paper is a deep review of the state of the art semi-supervised deep learning methods, focusing on methods dealing with the distribution mismatch setting. Under real world usage scenarios, a distribution mismatch might occur between the labelled and unlabelled datasets. Recent research has found an important performance degradation of the state of the art semi-supervised deep learning (SSDL) methods. Therefore, state of the art methodologies aim to increase the robustness of semi-supervised deep learning frameworks to this phenomena. In this work, we are the first to our knowledge to systematize and study recent approaches to robust SSDL under distribution mismatch scenarios. We think this work can add value to the literature around this subject, as it identifies the main tendencies surrounding it. Also we consider that our work encourages the community to draw the attention on this emerging subject, which we think is an important challenge to address in order to decrease the lab-to-real-world gap of deep learning methodologies.
\end{abstract}

}

\maketitle

\IEEEdisplaynontitleabstractindextext

%
\IEEEpeerreviewmaketitle

\ifCLASSOPTIONcompsoc
\IEEEraisesectionheading{\section{Introduction}\label{sec:introduction}}
\else
\section{Introduction}
\label{sec:introduction}

\IEEEPARstart{D}{eep} learning based approaches continue to provide more accurate results in a wide variety of  fields, from medicine to biodiversity
conservation \cite{calderon2018assessing,garcia2019convolutional,calderon2021real,icpr2020calderon,zamora2021enforcing,bermudez2020quality,oala2020ml4h,calvo2019assessing,calderon2020first}.
Most of deep learning architectures rely on the usage of extensively labelled
datasets to train models with  millions of parameters to estimate \cite{goodfellow2016deep,calderon2021correcting,calderon2021improving}.
Over-fitting is a frequent issue when implementing a deep learning
based solution trained with a small, or not representative dataset.  Such phenomena often causes   poor generalization performance during its real world usage. In spite of this risk, the acquisition of a sufficiently sized and representative sample, through rigorous procedures and standards, is a pending challenge, as argued in \cite{balki2019sample}.  Moreover, procedures to determine whether a dataset is large and/or representative enough is still an open subject in the literature, as discussed
in \cite{mendezusing}.

Often labels are expensive to generate, especially in  fields developed by highly trained medical professionals, such as radiologists, pathologists,
or psychologists \cite{calvo2019assessing,dwyer2018machine,calderon2018assessing,iglovikov2018paediatric}.
Examples of this include the labelling of hystopathological 
images, necessary for training a deep learning model for its usage
in clinical procedures \cite{calvo2019assessing}. Therefore, there is an increasing interest for dealing with scarce labelled data to feed deep learning architectures, stimulated by the success of deep learning based models \cite{oliver2018realistic}. 

Among the most popular and simple approaches to deal with limited
labelled observations and diminish model over-fitting is data augmentation. Data augmentation adds artificial observations to the training dataset, using simple transformations of real data samples; namely image rotation, flipping, artificial noise addition \cite{goodfellow2016deep}. A description of simple data augmentation procedures for deep learning architectures can be found
in \cite{yu2017deep}. More complex data augmentation techniques
make use of generative adversarial networks. Generative models  approximate the data
distribution, which can be sampled to create new observations, as
seen in \cite{shin2018medical,zhu2018emotion,frid2018synthetic}
with different applications. Data augmentation is implemented in popular deep learning frameworks, such as \emph{Pytorch}
and \emph{TensorFlow} \cite{paszke2017automatic}.

Transfer learning is also a common approach for dealing with the lack of enough
labels. It first trains a model $f$ with an external
or source-labelled dataset, hopefully from a similar domain. Secondly,  parameters are fine-tuned with the intended, or target dataset \cite{torrey2010transfer}.
Similar to data augmentation,\emph{ TensorFlow } and \emph{Pytorch
} include the weights of  widely used deep learning models trained 
in general purpose datasets as ImageNet \cite{deng2009imagenet},
making its usage widespread. Its implementation yields better results
with more similar source and target datasets. A detailed review on
deep transfer learning can be found in \cite{tan2018survey}.

Another alternative to deal with small labelled datasets is \gls{SSDL} which enables  the model to take advantage of unlabelled or even  noisily-labelled data \cite{xiao2015learning,kong2019recycling}. As an application example, take the problem of training a face based apparent  emotion recognition model. Unlabelled videos and images of human faces  are available on the web, and can be fetched with a web crawler. Taking advantage of such unlabelled information might yield improved accuracy and generalization for deep learning architectures. 

One of the first works in the literature regarding semi-supervised learning is \cite{shahshahani1994effect};
where different methods for using unlabelled data were proposed. More recently, with the increasing development and usage of deep learning architectures, semi-supervised learning methods are attracting more attention. An important
number of \gls{SSDL} frameworks are general enough to allow the usage of popular deep learning architectures in different application domains \cite{oliver2018realistic}. Therefore, we argue that it is necessary to review and study the relationship between recent deep learning based semi-supervised techniques, in order to spot missing gaps and boost research in the field.  Some recent semi-supervised learning reviews are already available in the literature. These are detailed in Section \ref{subsec:Previous-work}.  Moreover, we argue that it is important to discuss the open challenges of implementing \gls{SSDL} in real-world settings, to narrow the lab-application gap. One of the remaining challenges is the frequent distribution mismatch between the labelled and unlabelled data, which can hinder the performance of the \gls{SSDL} framework.

\subsection{Previous work \label{subsec:Previous-work}}

In \cite{zhu2005semi}, an extensive review on semi-supervised approaches for machine learning was developed. The authors defined the following semi-supervised approaches: self-training, co-training, and graph based methods. However, no deep learning based concepts were popular by the time of the survey, as auto-encoder and generative adversarial networks were less used, given its high computational cost and its consequent impractical usage. 

Later, a review of semi-supervised learning methods was developed in \cite{pise2008survey}. In this work, authors enlist self-training, co-training, transductive support vector machines, multi-view learning and generative discriminative approaches. Still deep learning architectures were not popular by the time. Thus, semi-supervised architectures based on more traditional machine learning methods are reviewed in such work. 

A brief survey in semi-supervised learning for image analysis and
natural language processing was developed in \cite{prakash2014survey}. The study defines the following semi-supervised learning approaches: generative models, self-training, co-training, multi-view learning
and graph based models. This review, however, does not focus on deep semi-supervised learning approaches.

A more recent survey on semi-supervised learning for medical imaging can be found in \cite{cheplygina2019not}, with  different machine learning based approaches listed. Authors distinguished self-training, graph based, co-training, and manifold regularization approaches for semi-supervised learning. More medical imaging solutions based on transfer learning than semi-supervised learning were found by the authors, given its simplicity of implementation.

In \cite{oliver2018realistic}, authors experimented with some recent \gls{SSDL} architectures and included a short review. The authors argued that typical testing of semi-supervised techniques is not enough to measure its performance in real-world applications. For instance, common semi-supervised learning benchmarks do not include unlabelled datasets with observations from classes not defined in the labelled data. This is referred to as distractor classes or collective outliers \cite{singh2012outlier}. The authors also highlight  the lack of tests around the interaction of semi-supervised learning pipelines with other types of learning, namely transfer learning.

More recently, in \cite{van2020survey}, the authors extensively review different semi-supervised learning frameworks, mostly for deep learning architectures. A detailed concept framework around the key assumptions of most \gls{SSDL} (low density/clustering and the manifold assumptions) is developed. The taxonomy proposed for semi-supervised  methods include two major categories: inductive and transductive based methods. Inductive methods build a mathematical model or function that can be used for new points in the input space, while transductive methods do not. According to the authors, significantly more semi-supervised inductive methods can be found in the literature. These methods can be further categorized into: unsupervised pre-processing, wrapper based, and intrinsically semi-supervised methods \cite{van2020survey}. The authors mentioned the distribution mismatch challenge for semi-supervised learning introduced in \cite{oliver2018realistic}. However, no focus on techniques around this subject was done in their review. 

In \cite{schmarje2021survey}, a review of semi-, self- and unsupervised learning methods for image classification was developed. The authors focus on the common concepts used in these methods (most of them based on deep learning architectures). Concepts such as pretext or proxy task learning, data augmentation, contrastive optimization, etc., are described as common ideas within the three learning approaches. A useful set of tables describing the different semi-supervised learning approaches along with the common concepts is included in this work.  After reviewing the yielded results of recent semi-supervised methods, the authors conclude that few of them include benchmarks closer to real-world  (high resolution images, with similar features for each class). Also,   real-world settings, such as class imbalance and noisy labels are often missing.

We argue that a detailed survey in \gls{SSDL} is still missing, as
common short reviews included in \gls{SSDL} papers usually focus
on its closest related works. The most recent semi-supervised
learning surveys we have found are outdated and do not focus on deep learning based approaches. We argue that recent \gls{SSDL} approaches add new perspectives to the semi-supervised learning framework. However, more importantly, to narrow the lab-to-application gap, it is necessary to fully study the state of the art in the efforts to address such challenges. In the context of \gls{SSDL}, we consider that increasing the robustness of \gls{SSDL} methods to the distribution mismatch between the labelled and unlabelled datasets is key. Therefore, this review focuses on the distribution mismatch problem between the labelled and the unlabelled datasets.

In Section \ref{subsec:Semi-supervised-learning-framework}, a review of the main ideas of semi-supervised learning is carried out. Based on the concepts of both previous sections, in Section
\ref{sec:ssdl} we review the main approaches for \gls{SSDL}. Later we address the different methods developed so far in the literature regarding \gls{SSDL} when facing a distribution mismatch between $S^{(u)}$ and $S^{(l)}$. Finally, we discuss the pending challenges of \gls{SSDL} under distribution mismatch conditions in Section 
 \ref{sec:conclusion}.

\subsection{Semi-supervised   learning}\label{subsec:Semi-supervised-learning-framework}

In this section we  describe the key terminology to analyze \gls{SSDL} in more  detail. We based our terminology on the semi-supervised learning analytical framework developed in \cite{balcan200621}. This framework extends the learning framework
proposed in \cite{valiant1984theory}, as a machine learning theoretical framework.

A model $f_{\textbf{w}}$ is said to be semi-supervised, if it is trained using  a set of labelled observations $S^{(l)}$, along a set of unlabelled observations 
$S^{(u)}=\left\{ \mathbf{x}_{n_{1}},\mathbf{x}_{n_{2}},\ldots,\mathbf{x}_{n_{u}}\right\},\label{eq:UnlabelledDataset}$  with the total number of observations
$n=n_{l}+n_{u}$. Frequently, the number of unlabelled observations $n_{u}$ is considerably higher than the number of labelled observations. This makes $n_{u}\gg n_{l}$, as labels are expensive to obtain in different domains.  If the model $f_{\textbf{w}}$ corresponds to a \gls{DNN}, we refer to \gls{SSDL}. The deep model $f_{\textbf{w}}$ is often referred to as a back-bone model.  In semi-supervised learning,  additional information is extracted from an unlabelled  dataset $S^{(u)}$. Therefore,  training a deep model can be extended to $f_{\mathbf{w}}=T\left(S^{(l)},S^{(u)},f_{\mathbf{w}}\right)$. The estimated hypothesis should classify
test data in $\mathbf{x}\in S^{(t)}$with a higher accuracy
than just using the labelled data $S^{(l)}$.

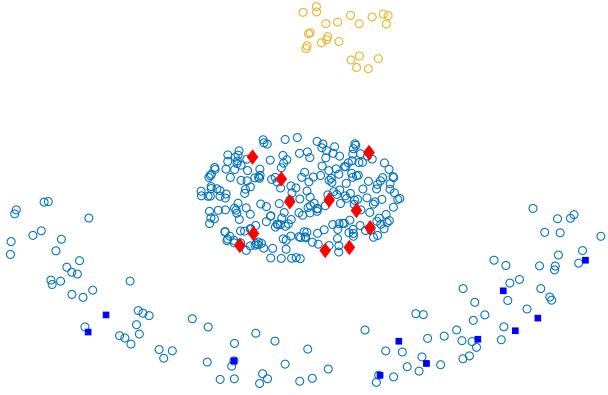
\begin{figure}
 \begin{tikzpicture}
\begin{axis}[  width=3.2in, height=2.0in, at={(2.08in,0.863in)}, scale only axis, xmin=-15, xmax=15, xtick={\empty}, ymin=-15, ymax=15, ytick={\empty}, axis line style={draw=none}, ticks=none, axis x line*=bottom, axis y line*=left ] \addplot[only marks, mark=o, mark options={}, mark size=1.5000pt, draw=mycolor1] table[row sep=crcr]{  x	y\\ 1.34788434389858	-3.78769359877904\\ -2.51924837434558	-3.81217279732888\\ -0.97294423963072	-1.62233379390213\\ 1.86243207206122	3.24153547594621\\ 1.0058953928319	2.46319144138628\\ -4.27964546954531	-1.68774740746495\\ -1.72796959154099	-0.720048303254643\\ 1.90697749682142	-0.477409003699997\\ -0.793117362361178	-2.26238562594316\\ -3.18152884935167	-1.23791157135848\\ -0.10480420249429	3.46364958058214\\ -2.23503717987381	-3.62592618265162\\ -3.25614147524135	3.15484364203757\\ 1.157338717612	-0.654972810338471\\ 3.7202491150718	1.01841891224233\\ -4.42918874782873	1.81984500968561\\ 3.88857092293528	-2.11557496736529\\ -1.88986096941092	4.52950254121662\\ -3.62090566597146	3.33991889521602\\ -2.29600240695142	1.53928589699436\\ -4.15519428944562	0.677830583515632\\ 1.54165592141257	3.46864631504736\\ -0.753696882192922	-3.36284139875153\\ -0.0486169075123621	-3.15114754259574\\ 0.0497240032921816	-1.39839824804432\\ -2.06698759794562	1.50564304466344\\ 1.44816342284008	1.49686335694743\\ -4.0047566468377	0.537138032245073\\ 0.174749941382146	-3.10149189672082\\ -4.5170724831718	-2.05590029229268\\ -1.04669319483492	-2.13350068989428\\ -0.904544816453283	-3.25456778853217\\ -0.527859943113577	0.903053624314981\\ 4.33920560515153	1.07392366429404\\ 3.6776586488108	0.664445544041692\\ -1.54120503899035	2.92721581530349\\ -1.20688532682497	-2.09344399598313\\ 0.356237358625647	1.21482552472793\\ -3.49652707590645	1.56732041119586\\ 3.72908207280299	-2.99158497958802\\ 1.2253970760855	3.67913774135017\\ -1.51578974346619	-4.7609604979602\\ -2.97596026212996	2.53094676435635\\ -3.11869271430054	3.32280470583566\\ -1.15540026954327	-2.15088716633378\\ -3.71119794484696	2.21850018448178\\ 2.53964819762811	3.82558146890448\\ -3.76199373342047	-2.68181376248871\\ -4.91514263619082	0.12313448126514\\ -4.47771082802298	0.101861828194191\\ 2.32281011540893	-2.77119679834172\\ -4.45267915284205	-1.60537040713877\\ -2.5144397225413	0.831725931827728\\ 4.29103498275598	2.20693030335886\\ 3.60300367172791	-0.350631684549354\\ -3.12140519138815	-0.46026360737941\\ -2.79927587501444	-2.6484118454389\\ -1.42229979828612	-4.00282323255264\\ 4.81656391438836	-0.0828880920427327\\ -0.533081334138213	-3.03866330941464\\ 0.40853769295341	-1.22122958555207\\ -0.493197898571858	-4.78383682554886\\ 2.53910833297688	-2.83210181804022\\ 3.52493643104204	-3.13473770220358\\ -3.66647632788134	-3.39479529171261\\ -1.68423726602292	4.18174160033366\\ 2.36467010056948	-1.78682105057341\\ 2.20910234378741	-0.0750817456542388\\ 2.47813778231297	2.88257939991\\ -4.26661338451287	2.35592511593321\\ 1.8680888862117	0.581100154980238\\ 0.414675790905563	0.0207101747550428\\ 2.10581185947629	2.41991742308475\\ 3.51755038550365	-0.529191785008597\\ 4.48476393238402	1.62977792565516\\ -1.97173580038205	-3.61300803482958\\ 0.614104826077966	-0.459824583483704\\ -0.886117011434542	2.69066199222693\\ -1.85045514833622	4.29167698113273\\ -1.73272498403124	-1.90836816256318\\ 1.52648522730923	-2.13031857525513\\ -4.43917925729939	0.854439011252188\\ -0.918110912159635	3.29753064600284\\ 1.68992089804749	2.24487280587974\\ -1.29645080206911	1.57212180406382\\ 0.121069501060307	-2.70335754349021\\ -2.52055240620271	-2.45565367291851\\ -0.129035310951554	-3.08428920059003\\ 4.14551703750697	-1.33611018999343\\ -3.0690148130378	3.66020908605912\\ 1.72581637625216	0.235592344533761\\ 0.00390484848383614	1.57798101954334\\ 0.784437541002485	-0.860738882728722\\ -0.0682864042668215	-4.81650919868202\\ 1.28892780007083	0.199824500834597\\ -4.93472053319861	0.490199951348566\\ -4.24584185371665	2.20479625203453\\ -3.74275211934261	-0.0448745679529927\\ -3.16955906183064	2.6579216663686\\ -3.05837342998162	-1.73239654481233\\ -3.21819226734937	-1.00327310517887\\ -4.44009572671205	0.692934011947466\\ 2.57910144107984	-0.544027121240037\\ 4.71544939709778	-0.158672758142724\\ 0.78083194375003	-2.82320063075561\\ -1.47406438500097	1.41556005135291\\ -0.0882618119701891	-0.120204482340854\\ -3.75646010065912	-2.71340317367757\\ 4.34008939328369	0.638668055146177\\ -2.75245822591137	0.642035895785139\\ 2.81130927958473	2.74217551297622\\ -2.67383813930617	1.30254264829696\\ 1.185541328663	3.10656411475624\\ -1.00036372027286	-4.79563676838978\\ -0.281263994526372	0.743014501089304\\ 1.48575760020292	1.17235416479958\\ 1.76013587673776	-3.22933692449158\\ -2.65216823327326	2.1107165367182\\ 4.30507540966279	-1.82559677294193\\ -1.51227143428185	-1.44583604173478\\ 3.00654355534904	0.659041845710686\\ -2.99839416978642	-2.63729532625027\\ 2.09602292560651	-2.04381045689313\\ -3.30458904872286	-3.57178767318457\\ 1.61698534347308	3.98049907468626\\ 4.50530844474491	-0.76894541380446\\ -4.5499480350567	1.56609396976392\\ 2.29074367459614	2.72054623536133\\ -0.646813777654488	-2.07790185384304\\ 2.15910796604321	2.41144653451439\\ -2.12220969913028	-3.72422718642439\\ 0.28595495593992	3.57485721427379\\ 2.59854016205646	-0.225510490350977\\ 0.75732732544007	4.40173617790437\\ -0.474371356788215	0.357452280322869\\ 0.526401757958036	-3.52248897333209\\ 1.77710351096129	-2.51441593096205\\ -0.808237411951932	4.55219986809357\\ 2.52083785186664	-2.51368193758121\\ -4.48331452713105	1.71174560216715\\ -0.577186265216891	-0.202852584813628\\ 0.262541456103475	0.500083499203395\\ -2.0388859933668	1.73447026024994\\ 1.34851764863148	1.3465985336983\\ -0.761157999605823	-4.07553491297635\\ 3.93698127603328	-0.15738075892528\\ -2.69028700036253	2.99980966057576\\ 1.33512159511594	2.10493999372682\\ -2.99276258109198	-3.57552042399893\\ 0.814843195877366	-3.67687046067144\\ -4.52213712451656	-1.09819317528587\\ 4.517987456955	1.5583793279328\\ 2.7836273412086	1.74464408543831\\ 0.759502375851019	-1.09485921509223\\ 0.574538637549839	1.59849743944373\\ 4.55396462384766	0.35505317623376\\ -1.20003891558364	1.20381885427491\\ 2.74739772885327	-0.158234590149117\\ -0.99829125324573	2.33021466496132\\ 1.04230926737503	4.19421708086326\\ -2.48959190148252	-2.4796744007233\\ 3.2054825785333	-0.0317885181244904\\ 0.927493840426638	1.71901063361638\\ 2.50470564788774	1.57699485225677\\ 2.51061677046292	3.39504474033008\\ -1.0325955305037	0.740578564996807\\ -3.6256417706661	-3.2747653914414\\ 4.04674535788282	-2.46749749902938\\ 0.256405566574334	-2.73350445353266\\ 2.61546002498832	4.20824731112512\\ 1.82800657494138	-3.79189005567117\\ -1.09620951373721	-0.487444865191711\\ -2.51858823278206	-1.32517902855842\\ 3.86536544413575	1.30215312769615\\ -3.51870288227929	-3.05812144647702\\ -2.12269460152083	-3.50675413867829\\ 0.434449753872404	-0.657732681159209\\ -3.62391745379851	2.82622755216444\\ 0.147266712266287	1.97581123838228\\ -2.97226370168057	1.33771498828658\\ 3.13475386730612	-2.60148695380651\\ 0.993822153419428	3.89891808185532\\ 2.86852560180514	-2.20811697143999\\ 4.19842821131306	-1.98128797295562\\ 4.06394834375874	2.70904576483016\\ -3.3214287101191	-3.37694769001958\\ -0.130044623611804	-1.1971000203407\\ -2.7175900840265	-4.1635259821461\\ 3.27909198093083	1.26651795168298\\ -2.11204549805273	1.58703966433759\\ 2.69272490368	1.69731560430857\\ 3.34331326232958	-1.18152542164868\\ 2.90638335996841	-0.981424382394812\\ 3.69041939152994	-1.87883924329953\\ -2.04958981646304	2.21759168804951\\ 3.97467207317219	1.55404167581177\\ -1.52265787742712	-1.4747874950464\\ 0.196264720114908	-3.17167170665396\\ 3.80985202772008	-1.62534373769924\\ -0.724244800748832	2.96144299556289\\ 0.31610680092152	-0.798431069289497\\ 3.23187453771025	3.24448483621346\\ -3.28641249161246	2.18620395176998\\ 2.02447340824766	-2.03728850908399\\ -0.511701876126907	-1.73602383167858\\ 1.10480024199157	-2.54294578405626\\ -3.29380701925784	-0.850950109774858\\ 0.400931920957017	3.12923751348493\\ 2.54103407368454	-3.05267986315118\\ 3.39791005730366	-2.55992207863188\\ 2.39218395288646	0.315214935492278\\ -0.854251234910295	-1.18474309219452\\ 3.46423848598281	3.00692112593375\\ 2.65421336122848	4.09364792360299\\ -3.7058865362114	0.263001766319904\\ -0.265423952663378	-4.73090680495858\\ -1.28724810836324	-2.32927167565771\\ -3.73927238549458	-0.97808409159645\\ 2.14141095483845	0.118129459829551\\ -1.86320948359044	-3.98594254425744\\ 0.766836706371341	-0.90264778374077\\ -2.72053131764262	-0.823163502228583\\ -2.00641126645015	-0.518842707664534\\ 1.82711672698495	1.81648527455328\\ -2.25637668644807	-2.20342663651172\\ -4.0867540888118	-1.01574458317831\\ -0.214052926530936	4.70258263663164\\ -4.00532293369556	0.0877020186285816\\ -2.28109163684889	-3.75512559337537\\ 1.81894938621085	-4.29148075771366\\ -4.58979394717625	0.0896672323468691\\ 2.88753723415084	3.43368295850233\\ 2.18746034099824	1.10085029617677\\ -3.14947397193446	2.86698771618766\\ 1.57770087282113	0.451899571371698\\ 2.45330664634471	1.89387122770081\\ 2.98481017902705	2.74934714953843\\ 1.83667595647632	-2.06519214250887\\ -2.79344941483338	0.316662258503288\\ 3.30500639166496	3.52484101037277\\ -1.67368951993849	-14.2527122971697\\ -8.6727956282244	-11.0459613663112\\ 11.6735554789469	-5.38007832560646\\ -6.99204778047252	-11.9538233474954\\ -13.1857421825049	-2.98558483359394\\ 9.79761650238396	-11.182155223115\\ -12.636158374143	-0.377187769525827\\ -1.91626457437221	-13.8536368741975\\ -12.7756478827203	-2.62827860392181\\ -12.3020959043645	-6.49391043239645\\ -11.5221488121188	-5.48214774520602\\ 9.92210450444465	-10.1688312289566\\ 6.99692228170929	-10.9027676291967\\ -12.4298027089814	-0.325816985945432\\ 12.1747545831571	-7.82842717276645\\ -10.8976395200834	-4.97194899690153\\ -5.36446591179088	-9.5310940369387\\ 5.60663914428368	-9.13636118246352\\ 13.1917455343415	-1.6420318165132\\ -1.30846103045412	-11.289575329671\\ -10.7256254464826	-7.85245523934095\\ 10.216400473848	-6.72996808424904\\ -4.63220212059157	-12.9394859534834\\ -3.30262080045398	-14.2531493986244\\ 8.36504397938202	-11.3264524248556\\ -6.76716426540106	-12.6367713347877\\ 8.57982889918797	-11.7790893307918\\ -2.06203508384378	-14.6224619454252\\ 11.3526772889599	-0.869522949281428\\ 11.9291577536051	-2.74529055190578\\ 12.4435262333691	-5.39295675393273\\ 3.11160353935822	-10.4201819737228\\ -3.99798557156099	-14.3007298299288\\ -11.2799258416339	-5.88470450347078\\ 5.17947538761549	-13.3002767885611\\ -7.78711999978881	-9.09738215233632\\ 11.736633900904	-7.16457499850258\\ 3.78071049058391	-13.9687028378036\\ 3.66745681100337	-14.5342780171866\\ 13.5964279979805	-5.16332087544049\\ -0.808887124827149	-13.0989724657858\\ 12.5585135568363	-1.67772380930445\\ -11.2718603975706	-7.62491914713791\\ -8.10233461965448	-9.99970552372985\\ -12.0549299721154	-4.19273225112835\\ -4.58757779853886	-10.1839256133022\\ 6.82085677314921	-13.145554761509\\ 0.300364743618776	-11.9170353540082\\ 8.99009355656915	-9.22968776833356\\ 11.0593626212675	-8.76624014228668\\ 12.2666463208469	-8.07462876551583\\ 13.3649808653502	-1.35535890567824\\ -14.0017719937425	-0.979813895495872\\ 7.91499960075289	-12.6809317880522\\ 1.30818992244584	-13.4825880083282\\ 8.42057494028244	-9.66168729718113\\ -14.2875407854671	-4.48495733029704\\ -0.0926203334942355	-14.4467041093387\\ -3.29315457990293	-11.4721708582036\\ -8.01566208100697	-8.89477011976285\\ 10.6883011772161	-6.44706155593881\\ -3.43218761184453	-13.2633087534323\\ 12.4110374933933	-5.69390728518195\\ -14.0813459609219	-1.29093426512876\\ 4.5117713603433	-14.1011238485517\\ -8.9342182074951	-10.8679779945012\\ -11.005009476059	-6.02085158437781\\ 12.6854897700323	-2.78182658225884\\ -14.2550233231504	-3.47625923550945\\ 12.5985295063657	-4.52966501287189\\ -2.24927765058294	-10.6721366468829\\ -10.6325854086674	-10.177939144891\\ 5.97208036531219	-9.21063024632165\\ 5.76193962411199	-13.6498656454944\\ -7.96325934424891	-10.7358082859461\\ 7.60745017023558	-10.4117481488768\\ 4.93902549946545	-12.1452342461829\\ 9.44118742430412	-4.93506846496422\\ 7.86755774392234	-11.2553562295145\\ 14.6831906502779	-3.05781756856589\\ 7.92205324840739	-7.16464545376027\\ 8.49738336945106	-8.2417202489313\\ -11.8289762191843	-6.57975259490858\\ -10.2517473208361	-7.2880696605865\\ -6.34539293064611	-12.0565568604102\\ 10.0223382469757	-5.34560994090095\\ 5.90326674980102	-12.5339878003633\\ 6.1813932787359	-11.0773444163436\\ -11.7906564455857	-3.28470369344268\\ 13.7836727867888	-4.17642600817796\\ -10.4395046015157	-1.6227123560413\\ -8.37509013577598	-11.5267573182143\\ 4.1270617353251	-12.062773106688\\ -8.4168527479887	-6.5748531094727\\ 10.1151433127125	-8.09606044126017\\ 0.523907478509714	-14.2122589641928\\ 8.23483887035797	-12.4542245257963\\ -12.2430231902068	-6.8424498623545\\ -7.4841676397256	-9.2890381808974\\ -3.32101319365981	-12.8420806966966\\ }; \addplot[only marks, mark=diamond*, mark options={}, mark size=2.5883pt, color=red, fill=red] table[row sep=crcr]{  x	y\\ 3.30500639166496	3.52484101037277\\ -2.40770893687743	3.1725315530004\\ -0.582032915286247	-0.341458382661029\\ -0.995260345187223	1.4513933569899\\ 3.35995227707087	-2.38948220792861\\ 1.35135098584693	-0.205673563026516\\ 1.15760606508656	-4.16915262779236\\ -2.36344402840537	-2.83979301059485\\ 2.34096424216446	-3.93128249935331\\ 2.68860616291416	-1.02517812676055\\ -3.0292681183698	-3.77668428780313\\ }; \addplot[only marks, mark=o, mark options={}, mark size=1.5000pt, draw=mycolor2] table[row sep=crcr]{  x	y\\ 2.83862312704087	11.1104835885437\\ 0.972501059357381	12.1450281683616\\ 0.419285162461969	12.9246795987271\\ 2.69422070118195	10.2039507374041\\ 3.46138655270864	14.1697603448793\\ 0.335007267105099	12.8669402394011\\ 0.219611532070311	11.6937126133071\\ 0.733941684769282	14.5709245193486\\ 4.15547556657229	13.6222443015899\\ 1.83139676333078	12.2485741521286\\ 1.76655402781827	13.7758448871333\\ 2.42405564501876	10.7980911198727\\ 0.725712393848257	14.989741086542\\ 2.82457478408879	13.6446351427372\\ 4.24264212294233	14.2818682244769\\ 1.22755826685916	12.3848090620048\\ 0.0680573360232656	14.5421479882028\\ 1.28148224672195	12.6284301534542\\ 3.271635956225	10.1059567954746\\ 0.26818688364885	11.9073278097909\\ 1.18716013450757	13.6536280767321\\ 4.00696874419892	14.4044698698927\\ 2.39849346794673	14.3003772108171\\ 3.76139719766781	10.8996794352713\\ }; \addplot[only marks, mark=square*, mark options={}, mark size=1.0607pt, color=blue, fill=blue] table[row sep=crcr]{  x	y\\ -3.32101319365981	-12.8420806966966\\ 6.12414278849823	-13.0471682160266\\ -9.59731053762446	-9.23532806970081\\ 4.7704227862358	-11.3017816073467\\ 3.84724892730685	-13.97474287544\\ 9.89333230261518	-7.3385568665649\\ 13.925249968935	-4.93511507008488\\ 10.4841680001104	-10.4774951103298\\ 11.5914713801182	-9.48434651279117\\ 8.64638485944899	-11.1455887930408\\ -10.4757085288824	-10.5783149809143\\ }; \end{axis} \end{tikzpicture} 

\caption{Semi-supervised setting, circles represent the unlabelled observations
$S^{(u)}$, the filled shapes correspond to labelled observations $S^{(l)}$
of $K=2$ classes, and the yellow circles correspond to unlabelled
observations or members of the distractor
class.\label{fig:Semi-supervised-setting,-circles}}

\end{figure}

 Figure \ref{fig:Semi-supervised-setting,-circles}
plots a semi-supervised setting with observations in $d=2$ dimensions.  The label can also correspond to an array, $\mathbf{y}_{i}\in\mathbb{R}^{k}$,
in case of using a $1-K$ encoding (one-hot vector) for classifying observations in
$K$ classes, or $y_{i}\in\mathbb{R}$ for regression. More specifically,
observations of both the labelled and unlabelled dataset belong to the observation space $\mathbf{x}_{i}\in\mathcal{X}$ and labels lie within the label space $\mathcal{Y}$. For instance, observation for binary images of written digits with $d$ pixels would make up for an observation space $\mathcal{X}\in\{0,1\}^{d}$, and its label
set is given as $\mathcal{Y}=\left\{ 0,1,\ldots,9\right\} $.

 The concept class $\mathcal{C}_{k}$ corresponds to all the valid combinations of values in the array $\mathbf{x}_{i}\in\mathbb{R}^{d}$
for a specific class $k$. For example, for the digit $1$, a subset of all possible of observations that belong to class $k$ belong to the concept $\mathcal{C}_{k}$.  The concept class models  all the possible images of the digit $1$. In such case  $\mathbf{x}_{i}\in\mathcal{C}_{k=1}$. The concept class $\mathcal{C}=\left\{ \mathcal{C}_{1},\ldots,\mathcal{C}_{k}\right\} $
includes all the possible observations which can be drawn for all
the existing classes in a given problem domain.

From a data distribution perspective, usually  the population density function of the concept class $p_{\mathbf{x}\sim\mathcal{C}}\left(\mathbf{x}\right)=p\left(\mathbf{x}|y=1,\ldots,K\right)$ and the density for each concept $p_{\mathbf{x}\sim\mathcal{C}_{k}}\left(\mathbf{x}\right)=p\left(\mathbf{x}|y=k\right)$
is unknown. Most semi-supervised methods assume that   both $S^{(u)}$
and labelled data $S^{(l)}$ sample the concept class density, making  $p_{\mathbf{x}\sim S^{(l)}}\left(\mathbf{x}\right)$
and $p_{\mathbf{x}\sim S^{(u)}}\left(\mathbf{x}\right)$ very similar \cite{van2020survey}.  A labelled and an unlabelled dataset, $S^{(l)}$ and $S^{(u)}$, respectively, are said to be identically and independently sampled if the density functions $p_{\mathbf{x}\sim S^{(u)}}$ and $p_{\mathbf{x}\sim S^{(l)}}$ are identical and are statistically independent.

 However, in real-world settings different violations to the  \gls{IID} assumption can be faced. 
For instance, unlabelled data is likely to contain observations
which might not belong to any of the $K$ classes. Potentially,  this could lead
to a different sampled density function from\textbf{ }$p_{\mathbf{x}\sim\mathcal{C}}\left(\mathbf{x}\right)$.
These observations belong to a distractor class dataset $\mathbf{x}\in \mathcal{D}$,
and are drawn from a theoretical distribution of the distractor class
$p_{\mathbf{x}\sim\mathcal{D}}\left(\mathbf{x}\right)$.
Figure \ref{fig:Semi-supervised-setting,-circles} shows distractor
observations drawn from a distractor distribution $p_{\mathbf{x}\sim\mathcal{D}}\left(\mathbf{x}\right)$
in yellow. 

A subset of unlabelled observations from $S^{(u)}$, referred to as ${S^{(u)}}_D$ are said to belong to a distractor class, if they are drawn from a different distribution than the observations that belong to the concept classes. The distractor class is frequently semantically different than the concept classes. 

Different causes for a violation to the \gls{IID} assumption for $S^{(u)}$ and $S^{(l)}$ might be faced in real-world settings.   These are enlisted as follows, and can be found with different degrees \cite{kairouz2019advances}:
\begin{itemize}
    
    \item Prior probability shift: The label distribution in the dataset $S^{(l)}$ might differ when compared to $S^{(u)}$.   A specific case would be the label imbalance of the labelled dataset $S^{(l)}$ and a balanced unlabelled dataset.
    \item Covariate shift: A difference in the feature distributions between the datasets  $S^{(l)}$ with respect to $S^{(u)}$ with the same classes in both, might be sampled, leading to a distribution mismatch. In a medical imaging application, for example, this can be related to the difference in the  distribution of the sampled features between $S^{(l)}$ and $S^{(u)}$. This can be caused by the difference of the patients sample.
  
    \item Concept shift: This is associated to a shift in the labels of $S^{(l)}$ with respect to $S^{(u)}$ of data with  the same features. For example, in the medical imaging domain, different practitioners might categorize the same x-ray image into different classes. This is very related to the   problem of noisy labelling \cite{frenay2013classification}.
    \item Unseen classes: The dataset $S^{(u)}$ contains observations of unseen or unrepresented classes in the dataset $S^{(l)}$. One or more distractor classes are sampled in the unlabelled dataset.  Therefore, a mismatch in the number of labels exist, along with a prior probability shift and a feature distribution mismatch. 
    
\end{itemize}

Figure \ref{fig:Semi-supervised-setting,-circles} illustrates a distribution mismatch setting. The circles correspond to unlabelled data and the squares and diamonds to the labelled dataset.  The labelled and unlabelled data for the two classes are clearly imbalanced and sample different feature values.   In this case, all the blue unlabelled observations are drawn from the concept classes. However, the yellow unlabelled observations, can be considered to have different feature value distributions. Many \gls{SSDL} methods make usage of   the clustered-data/low-density separation assumption together with  the manifold
hypothesis \cite{rifai2011manifold}.    

\section{Semi-supervised Deep Learning}\label{sec:ssdl}
In this section we study recent semi-supervised deep learning architectures. They are divided into different categories. Such categorization is meant to ease its analysis. However each category is not mutually exclusive with the rest of them, as there are several methods that mix concepts of two or more categories.    This serves as a background to understand current \gls{SSDL} approaches to deal with the distribution mismatch between $S^{(u)}$ and $S^{(l)}$.

\subsection{Pre-training for semi-supervised deep learning\label{subsec:Semi-supervised-pre-training-and}
}

A basic approach to leverage the information from an unlabelled dataset $S^{(u)}$, is to perform as a first step  an unsupervised pre-training
of the classifier $f_{\textbf{w}}$.  In this document we refer to it as \gls{PT-SSDL}. A straightforward way to implement \gls{PT-SSDL}, is to pre-train the encoding section of the model $h_{\textbf{w}_{\textrm{FE}}}^{\left(\textrm{FE}\right)}\left(\textbf{x}_{i}\right)$
to optimize a \emph{proxy }or \emph{pretext \cite{tran2019semi}
}task  $\delta$. The proxy task does not need the specific labels, allowing the usage of unlabelled data.  This proxy loss   is minimized during training, and enables the usage of unlabelled data:
\begin{equation}
\mathcal{\:L}_{u}^{\left(p\right)}\left(S^{(u)},\textbf{w}_{\textrm{FE}}\right)=\:\sum_{\textbf{x}_{i}\in S^{(u)}}\delta\left(r_{i},f_{\textrm{proxy}}\left(f_{\textbf{w}_{\textrm{FE}}}\left(\Psi^{\eta}\left(\textbf{x}_{i}\right)\right)\right)\right),
\label{eq:UnsupervisedLoss}
\end{equation}

where the function $\delta$ compares the proxy label $r_i$ with the output of the proxy model $f_{\textrm{proxy}}$. The proxy task can be optimized also using labelled data. The process of optimizing a proxy task is also known as self-supervision \cite{jing2020self}. This can be done in a pre-training step or during training, as seen in the models with unsupervised regularization.  A simple approach for this \emph{proxy }or\emph{ auxiliary } loss is to minimize the unsupervised reconstruction error. This is similar to the usage of a consistency function $\delta$, where the proxy task corresponds to reconstruct the input, making 
$\delta\left(\textbf{x}_{i},h_{\textbf{w}_{\textrm{DE}}}^{\textrm{(DE)}}\left(h_{\textbf{w}_{\textrm{FE}}}^{\left(\textrm{FE}\right)}\left(\textbf{x}_{i}\right)\right)\right) $. The usage of an auto-encoder based reconstruction means it is usually necessary to add a decoder path $h_{\textbf{w}_{\textrm{DE}}}^{\textrm{(DE)}}$, which is later discarded at evaluation time.  Pre-training can be performed for the whole
model, or in a per-layer fashion, as initially explored in  \cite{bengio2007greedy}. Moreover, pre-training can be easily combined  with other semi-supervised
techniques, as seen in \cite{lee2017deep}. 

In \cite{doersch2015unsupervised}, a \gls{CNN}
is pre-trained with image patches from unlabelled data, with the proxy
task of predicting the position of a new second patch. The approach
was tested in object detection benchmarks. In \cite{caron2019unsupervised} an unsupervised pre-training approach was proposed. It implements a proxy task optimization followed by a clustering step, both using unlabelled data. The proxy task consists of the random rotation of the unlabelled data, and the prediction of its rotation. The proposed method was tested against other unsupervised pre-training methods, using the  PASCAL Visual Object Classes 2007 dataset.

The proxy or auxiliary
task is implemented in different manners in \gls{SSDL}, as it is not exclusive to pre-training methods. This can be seen  in consistency
based regularization techniques, later discussed in this work. For instance in \cite{zhai2019s4l}, an extensive set of proxy tasks are added as an unsupervised regularization term, and compared with some popular regularized \gls{SSDL} methods. The authors used the \gls{ILSVRC} for the executed benchmarks. The proposed method  showed a slight accuracy gain, with no statistical significance, against other two unsupervised regularization based methods.

\subsection{Pseudo-label semi-supervised deep learning\label{subsec:Pseudo-label-semi-supervised-lea}}

In \gls{PLT-SSDL} or also known as self-training, self-teaching or bootstrapping, pseudo-labels are estimated
for unlabelled data, and used for  model fine-tuning.  A straightforward approach of pseudo-label based training consisting in co-training two models can be found in  \cite{balcan200621}. 

In co-training, two or more different sets of input dimensions or \textit{views}  are used to train two or more different models. Such views can be just the result of splitting the original input array $\textbf{x}_i$.  For instance, in two-views $v_1$ and $v_2$ co-training \cite{balcan200621},  two labelled datasets  $S^{\left(l,v_{1}\right)}$ and $S^{\left(l,v_{2}\right)}$ are used. In an initial iteration $i=1$, two models are trained using the labelled dataset, yielding the two view models $\widetilde{\textbf{w}}_{i}^{\left(v_{1}\right)}=T\left(f_{\textbf{w}},S^{\left(l,v_{1}\right)}\right)$ and $\widetilde{\textbf{w}}_{i}^{\left(v_{1}\right)}=T\left(f_{\textbf{w}},S^{\left(l,v_{2}\right)}\right)$.   This can be considered as a pre-training step.  The resulting  models can be referred to as an ensemble of models $\textbf{f}_{\widetilde{\textbf{w}}_{1}}=\left[f_{\widetilde{\textbf{w}}_{1}^{\left(v_{2}\right)}},f_{\widetilde{\textbf{w}}_{2}^{\left(v_{2}\right)}}\right]$. 

As a second step, the disagreement probability $\mathbf{Pr}_{\textbf{x}_{i}\sim S^{(u)}}\left[f_{\widetilde{\textbf{w}}_{1}^{\left(v_{2}\right)}}\left(\textbf{x}_{j}\right)\neq f_{\widetilde{\textbf{w}}_{2}^{\left(v_{2}\right)}}\left(\textbf{x}_{j}\right)\right]$ with $\textbf{x}_{j} \in S^{(u)}$  is   used to estimate new labels or pseudo-labels $\widehat{y}_{j}^{(i,k)}=f_{\widetilde{\textbf{w}}_{i}^{\left(v_{k}\right)}}\left(\textbf{x}_{j}\right)$. The final pseudo-labels for each observation $\textbf{x}_{j}$ can be the result of applying a view-wise summarizing operation $\mu$ (like averaging or taking the maximum logits) making $\widehat{y}_{j}^{(i)}=\mu\left(\textbf{f}_{\textbf{w}_{i}}\left(\textbf{x}_{j}\right)\right)$.  

The set of pseudo labels for the iteration i can be represented as $\widehat{S}_{i}=\mu\left(\textbf{f}_{\textbf{w}_{i}}\left(S^{(u)}\right)\right)$. In co-training \cite{balcan200621}, the agreed observations for the two models are picked in the function $\widetilde{S}_{i}=\varphi\left(\widehat{S}_{i}\right)$, as highly confident observations. The pseudo-labelled data with high confidence are included in the labelled dataset $S_{i+1}^{\left(r\right)}=S^{(l)}\bigcup\widetilde{S}_{i}$ as pseudo-labelled observations.  Later the model is re-trained for $i=2,..,\vartheta$ iterations repeating the process of pseudo-labelling, filtering the most confident pseudo-labels and re-training the model. In general, we refer to a pseudo-labelling to the idea of estimating the hard labels $\widehat{y}_{j}^{(i)}$.

In \cite{dong2018tri} the \gls{Tri-Net} was proposed. Here an ensemble $\textbf{f}_{\textbf{w}}$ of \gls{DCNN}s is trained with $k=1,2,3$ different top models, with also $k=1,2,3$ different labelled datasets $S_{i}^{(l,k)}$. The output posterior probability is the result of the three models voting. This results in the pseudo-labelled for the whole evaluated dataset $\widetilde{S}_{i}$, with $i=1$ for the first iteration. The pseudo-label filtering operation $\varphi$ includes the observations where at least two of the models agreed are included into the labelled dataset. The process is repeated for a fixed number of iterations. Also \gls{Tri-Net} can be combined with any regularized \gls{SSDL} approach. This combination was tested in \cite{sun2017enhancing}, and is referred to in this document as \gls{TriNet+Pi}.  In \cite{sun2017enhancing}, a similar ensemble-based pseudo-labelling approach is found.   In such work, a mammogram image classifier was implemented, with an ensemble
of classifiers that vote for the unlabelled observations. The observations with the highest confidence are added to the dataset, in an
iterative fashion.

Another recent deep self-training approach can be found in \cite{cicek2018saas}, named as \gls{SaaSM}. In a first step, the pseudo-labels are estimated by measuring the learning speed in epochs, optimizing the estimated labels as a probability density function $\widetilde{S}_{1}=f_{\textbf{w}_{1}}\left(S^{(u)}\right)$ with a stochastic gradient descent approach. The estimated labels are used to optimize an unsupervised regularized loss. \gls{SaaSM} was tested using the \gls{CIFAR-10} and \gls{SVHN} datasets. It yielded slightly higher accuracy to other consistency regularized methods such as mean teacher, according to the reported results. No statistical significance analysis was done.

\subsection{Regularized semi-supervised learning \label{sec:Regularized-semi-supervised-lear}}

In \gls{R-SSDL}, or co-regularized learning as defined in \cite{zhu2005semi},
the loss function of a deep learning model $f_{\textbf{w}}$
includes a regularization term using unlabelled data $S^{(u)}$: 
\begin{equation}
\overset{\underset{\textbf{w}}{\textrm{argmin}}\mathcal{L}\left(S,f_{\textbf{w}}\right)=}{\underset{\textbf{w}}{\textrm{argmin}}\sum_{\left(\textbf{x}_{i},y_{i}\right)\in S^{(l)}}\mathcal{L}_{l}\left(f_{\textbf{w}}\left(\textbf{x}_{i}\right),y_{i}\right)+\gamma\sum_{\textbf{x}_{j}\in S^{(u)}}\mathcal{L}_{u}\left(f_{\textbf{w}},\textbf{x}_{i}\right)}.\label{eq:reguLearning}
\end{equation}

The unsupervised loss $\mathcal{L}_{u}$ regularizes the model $f_{\textbf{w}}$ using the unlabelled observations $\textbf{x}_{j}$. The unsupervised regularization coefficient $\gamma$ controls the unsupervised regularization influence during the model training. We consider it an \gls{SSDL} sub-category, as a wide number of approaches have been developed inspired by this idea.  In the literature, different approaches for implementing the unsupervised
loss function $\mathcal{L}_{u}$ can be found. Sections \ref{subsec:Consistency-based-loss},
\ref{subsec:Augmented-regularized-loss} and \ref{subsec:Graph-based-regularized}
group the most common approaches for implementing it. 

\subsubsection{Consistency based regularization}\label{subsec:Consistency-based-loss}

A \gls{RC-SSDL} loss function measures how \emph{robust
} a model is, when classifying unlabelled observations in $S^{(u)}$ with different transformations applied to the unlabelled data. Such transformations usually perturb the unlabelled data without changing its semantics and class label (label preserving transformations).  For instance, in \cite{balcan200621}
 consistency assumption  $\chi^{\textrm{(CL)}}$ is enforced for two views in \cite{balcan200621}, using  the Euclidean distance:
$\delta\left(\textbf{x}_{j},f_{\textbf{w}}\right)=\left\Vert f_{\textbf{w}'}\left(\Psi^{\eta'}\left(\textbf{x}_{j}\right)\right)-f_{\textbf{w}}\left(\Psi^{\eta}\left(\textbf{x}_{j}\right)\right)\right\Vert.
$

Where  $\delta\left(\textbf{x}_{i},f_{\textbf{w}}\right)$
is the consistency function. Consistency  
can also be measured for labelled observations in $\textbf{x}_{j}\in S^{(l)}$. A number of \gls{SSDL} techniques are based on consistency regularization. Therefore we refer to this category as \gls{CR-SSDL}.

 A simple interpretation of the consistency regularization term  is the
increase of a model's robustness to noise, by using the data in $S^{(u)}$.
A consistent model output for corrupted observations implies a more
robust model, with better generalization.  Consistency can be measured between two deep learning models $f_{\textbf{w}'}$
and $f_{\textbf{w}}$ fed with two different views or random
modifications of the same observation, $\Psi^{\eta'}\left(\textbf{x}_{j}\right)$
and $\Psi^{\eta}\left(\textbf{x}_{j}\right)$. For this reason, some authors refer to consistency based \gls{SSDL} approaches as self-ensemble learning models \cite{luo2018smooth}. The consistency of two or more variations of the model is evaluated, measuring the overall model robustness. 

Consistency regularized methods  are based on the consistency assumption $\chi^{(\textrm{CL})}$. Thus, such methods can be related to other previously \gls{SSDL} approaches that also exploit this assumption. For instance, as previously mentioned, the consistency assumption is  also implemented in \gls{PT-SSDL} as the proxy task, where the model is pre-trained to minimize a proxy function. The consistency function can be thought as a particular case of the
aforementioned confidence function for self-training $\varphi$, using two or more views of the observations, as developed in \cite{balcan200621}. However, in this case the different corrupted views of the observation $\textbf{x}_{j}$
are highly correlated, in spite of the original co-training approach
developed in \cite{balcan200621}. Nevertheless, recent regularized \gls{SSDL}
models \cite{bachman2014learning,tarvainen2017mean,laine2016temporal}
simplify this assumption. They consider as a view of an observation $\textbf{x}_{j}$ its corruption
with random noise $\eta$, making up a corrupted view $\Psi^{\eta}\left(\textbf{x}_{j}\right)$.
The   independence assumption \cite{balcan200621}
between the views of co-training, fits better when measuring
the consistence between different signal sources, as seen in \cite{lv2018bi}. In such work, different data sources are used for semi-supervised human activity recognition.

In \cite{bachman2014learning} the \gls{Pi-M} was proposed. The consistency
of the deep model with random noise injected to its weights (commonly referred to as dropout) is evaluated. The weights $\textbf{w}'$ are a corrupted version of the \textit{parent} model with weights $\textbf{w}$, making up what the authors refer as a \textit{pseudo-ensemble}. The \gls{Pi-M} model was tested in \cite{tarvainen2017mean} using the \gls{CIFAR-10} and \gls{SVHN} datasets. Intersected yielded results of \gls{Pi-M} with the rest of the discussed methods in this work can be found in Table \ref{tab:cifar-10_ssdl}.

A consistency evaluation of both unlabelled and labelled datasets can be performed, as proposed in \cite{sajjadi2016regularization}, in the \gls{METM}. In such method, an unsupervised loss term for Transformation Supervision (TS) was proposed: $ \mathcal{L}_{u}^{\left(\textrm{TS}\right)}\left(f_{\textbf{w}},\textbf{x}_{i}\right)=\sum_{j}^{M}\sum_{k}^{M}\left\Vert f_{\textbf{w}}\left(\Psi_{j}\left(\textbf{x}_{i}\right)\right)-f_{\textbf{w}}\left(\Psi_{k}\left(\textbf{x}_{i}\right)\right)\right\Vert ^{2},$ where $M$ random transformations $\Psi$ are performed over the observation $\textbf{x}_{i}$. This can be used for unsupervised pre-training. Such loss term can be regarded as a consistency measurement. Furthermore, a \gls{ME} based loss function is used. It encourages non-overlapping predictions of the model. The \gls{ME} loss term is depicted as $\mathcal{L}_{u}^{\left(\textrm{ME}\right)}=\left\Vert -\prod_{k}^{K}f\left(\textbf{x}_{i}\right)\prod_{k}^{K}\left(1-f_{\textbf{w}}\left(\textbf{x}_{i}\right)\right)\right\Vert ^{2}$. The final unsupervised loss is implemented as $\mathcal{L}_{u}=\lambda_{1}\mathcal{L}_{u}^{\left(\textrm{ME}\right)}+\lambda_{2}\mathcal{L}_{u}^{\left(\textrm{TS}\right)}$, with the weighting coefficients $\lambda_{1}$ and $\lambda_{2}$ for each unsupervised loss term. \gls{METM} was tested with the \gls{SVHN} and \gls{CIFAR-10} datasets. Comparable results with the rest of reviewed methods in this work are depicted in Table   \ref{tab:cifar-10_ssdl}.

Later, authors in \cite{laine2016temporal} proposed the \gls{TEM}, which calculates the consistency of the trained model with the the moving weighted average of the predictions from different models along each training epoch $\tau$: \begin{equation} f_{\textbf{w}'_{\tau}}\left(\Psi^{\eta}\left(\textbf{x}_{i}\right)\right)=(1-\rho)f_{\textbf{w}_{\tau}}\left(\Psi^{\eta}\left(\textbf{x}_{i}\right)\right)+\rho f_{\textbf{w}_{\tau-1}}\left(\Psi^{\eta}\left(\textbf{x}_{i}\right)\right),\label{eq:TemporalEnsembling} \end{equation} with $\rho$ the decay parameter, and $\tau$ the current training epoch. The temporal ensemble evaluates the output of a temporally averaged model to a noisy observation $\Psi^{\eta}\left(\textbf{x}_{i}\right)$. This enforces temporal consistency of the model. Table \ref{tab:cifar-10_ssdl}  shows the yielded results by the \gls{TEM} method for the \gls{CIFAR-10} dataset.  Based on this approach, the authors in \cite{tan2019semi}, developed an \gls{SSDL} approach based on the Kullback-Leibler cross-entropy to measure model consistency. Different transformations $\Psi^{\eta}$ are applied to the input observations $\textbf{x}_{j}$. These correspond to image flipping, random contrast adjustment, rotation and cropping. The method was evaluated in a real world scenario with ultrasound fetal images for anatomy classification.

An extension of the temporal ensembling idea was presented by the authors of \cite{tarvainen2017mean}, in the popular \gls{MTM}. Instead of averaging the predictions of models calculated in past epochs, the authors implemented an exponential weight average: $ \textbf{w}'_{\tau}=\rho\textbf{w}'_{\tau-1}+\left(1-\rho\right)\textbf{w}'_{\tau} \label{eq:MeanTeacher} $ for a training epoch $\tau$, with an exponential weighting coefficient $\rho$. Such exponentially averaged model with parameters $\textbf{w}'$ is referred to by the authors as the \textrm{teacher} model. For comparison purposes, the yielded results by \gls{MTM} using the \gls{CIFAR-10} dataset are depicted in Table \ref{tab:cifar-10_ssdl}.

More recently, authors in \cite{miyato2018virtual}, proposed the  \gls{VATM}. They implemented a  generative
adversarial network to inject adversarial perturbations $\eta$
into the labelled and unlabelled observations. Artificially generated observations are compared to the original unlabelled data.  This results in adversarial noise encouraging a more challenging consistency robustness. Furthermore, the authors also added a conditional entropy term, in order to make the model more confident when minimizing it. We refer to this variation  as \gls{VATM+EM}. Both \gls{VATM} and \gls{VATM+EM} were tested with the \gls{CIFAR-10} dataset, thus we include the comparable results with the rest of the reviewed methods in Table \ref{tab:cifar-10_ssdl}.

Another variation of the consistency function $\mathcal{L}_{u}$ was developed in \cite{chen2018semi} in what the authors referred to as a memory loss function. We refer to this as the \gls{MeM}. This memory loss is based on a memory module, consisting of an embedding $m_{i}=\left(\check{\textbf{x}}_{i},\hat{\textbf{y}}_{i}\right)$. It is composed of the features extracted $\check{\textbf{x}}_{i}=h_{_{\textbf{w}^{\left(\textrm{FE}\right)}}}^{\left(\textrm{FE}\right)}\left(\textbf{x}_{i}\right)$ by the deep learning model $f_{\textbf{w}}$, and the corresponding probability density function computed $\hat{\textbf{y}}_{i}=f_{\textbf{w}}\left(\textbf{x}_{i}\right)$ (with $\hat{\textbf{y}}$ the logits output), for a given observation $\textbf{x}_{i}$. The memory stores one embedding $k$ per class $m_{k}=\left(\check{\textbf{\ensuremath{\textbf{x}}}}_{k},\hat{\textbf{\ensuremath{\textbf{y}}}}_{k}\right)$, corresponding to the average embedding of all the observations within class $k$. Previous approaches like the temporal ensemble \cite{laine2016temporal} needed to store the output of past models for each observation. In the memory loss based approach of \cite{chen2018semi} this is avoided by only storing one average embedding per class. In the second step, the memory loss is computed as follows: $\mathcal{L}_{m}=H\left(\textbf{p}_{i}\right)+\max\left(\textbf{p}_{i}\right)\delta_{\textrm{KL}}\left(\textbf{p}_{i},\hat{\textbf{y}}_{i}\right),\label{eq:MemoryLoss}$ where $\textbf{p}_{i}$ is they key addressed probability, calculated as the closest embedding to $\textbf{x}_{i}$, and $\hat{\textbf{y}}_{i}$ is the model output for such observation. The factor $\max\left(\textbf{p}_{i}\right)$ is the highest value of the probability distribution $\textbf{p}_{i}$ and $H\left(\textbf{p}_{i}\right)$ is the entropy of the key addressed output distribution $\textbf{p}_{i}$. The factor $\delta_{\textrm{KL}}\left(\textbf{p}_{i},\hat{\textbf{y}}_{i}\right)$ is the Kullback-Leibler distance of the output for the observation $\textbf{x}_{i}$ and the recovery key address from the memory mapping. Comparable results to the rest of the reviewed methods for the \gls{MeM} method are shown in Table \ref{tab:cifar-10_ssdl}  for the \gls{CIFAR-10} dataset.

More recently, an \gls{SSDL} approach was proposed in \cite{shi2018transductive}. This is referred to as the \gls{TransM}. The authors implement a transductive learning approach. This means that the unknown labels $\widetilde{y}$ are also treated as variables, thus optimized along with the model parameters $\textbf{w}$. Therefore, the loss function implements a cross-entropy supervised loss: $\mathcal{L}_{l}\left(f_{\textbf{w}},\textbf{x}_{i},y_{i}\right)=r_{i}H_{\textrm{CE}}\left(f_{\textbf{w}}\left(\textbf{x}_{i}\right),y_{i}\right), $ with $r_{i}$ an element of the set $R=\left\{ r_{i}\right\} _{i=1}^{n_{l}+n_{u}}$, which indicates the label estimation confidence level for an observation $\textbf{x}_{i}$. Such confidence level coefficient makes the model more robust to outliers in both the labelled and unlabelled datasets. The confidence coefficient is calculated using a k-nearest neighbors approach from the labelled data, making use of the observation density assumption $\chi^{(\textrm{CL})}$. This means that the label estimated is of high confidence, if the observations lies in a high density space for the labelled data within the feature space. As \gls{DCNN}s are meant to be used by the model, the feature space is learned within the training process, making necessary to recalculate $R$ at each training step $\tau$. As for the unlabelled regularization term:
$ \mathcal{L}_{u}\left(f_{\textbf{w}},\textbf{x}_{j}\right)=\lambda_{\textrm{RF}}\mathcal{L}_{\textrm{RF}}\left(f_{\textbf{w}},\textbf{x}_{j}\right)+\lambda_{\textrm{MMF}}\sum_{\textbf{x}_{i}}\mathcal{L}_{\textrm{MMF}}\left(f_{\textbf{w}},\textbf{x}_{j},\textbf{x}_{i}\right),\label{eq:transductive} $ it is composed of a robust feature measurement $\mathcal{L}_{\textrm{RF}}$ and a min-max separation term $\mathcal{L}_{\textrm{MMF}}$, where $\lambda_{\textrm{RF}}$ and $\lambda_{\textrm{MMF}}$ weigh their contribution to the unsupervised signal. The first term measures the feature consistency, thus using the output of the learned feature extractor of the model $\check{\textbf{x}}_{i}=h_{\textbf{w}_{\textrm{FE}}}^{\left(\textrm{FE}\right)}\left(\textbf{x}_{i}\right)$. The consistency of the learned features is measured with the Euclidian distance $\mathcal{L}_{\textrm{RF}}\left(\textbf{w},\textbf{x}_{j}\right)=\left\Vert h_{\textbf{w}_{\textrm{FE}}}^{\left(\textrm{FE}\right)}\left(\Psi^{\eta}\left(\textbf{x}_{j}\right)\right)-h_{\textbf{w}_{\textrm{FE}}}^{\left(\textrm{FE}\right)}\left(\Psi^{\eta'}\left(\textbf{x}_{j}\right)\right)\right\Vert ^{2}$. Regarding the second term, referred to as the min-max separation function, it is meant to maximize the distance between observations of different classes by a minimum margin $\rho$, and to minimize the distance from observations within the same class. It is implemented as follows: \begin{equation} \mathcal{L}_{\textrm{MMF}}\left(f_{\textbf{w}},\textbf{x}_{i},\textbf{x}_{j}\right)=r_{i}r_{j}\left\Vert f_{\textbf{w}}\left(\textbf{x}_{i}\right)-f_{\textbf{w}}\left(\textbf{x}_{j}\right)\right\Vert ^{2}\delta\left(\widetilde{y}_{i,}\widetilde{y}_{j}\right) \end{equation}  \[ -\textrm{min}\left(\left\Vert f_{\textbf{w}}\left(\textbf{x}_{i}\right)-f_{\textbf{w}}\left(\textbf{x}_{j}\right)\right\Vert ^{2}-\rho,0\right)\left(1-\delta\left(\widetilde{y}_{i,}\widetilde{y}_{j}\right)\right). \] With $\delta\left(\widetilde{y}_{i,}\widetilde{y}_{j}\right) = 1$ when $\widetilde{y}_{i} = \widetilde{y}_{j}$, or cero otherwise.  The first term in $\mathcal{L}_{\textrm{MMF}}$ minimizes the intra-class distance and the second term maximizes the inter-class observation distance. We highlight the theoretical outlier robustness of the method by implementing the confidence coefficient $r_{i}$. This coefficient is able to give lower relevance to unconfident estimations.  However, it is yet to be fully proved, as the experiments conducted in \cite{shi2018transductive} have not tested the model robustness to unlabelled single and collective outliers. The approach was also combined and tested against other consistency regularization approaches, like the \gls{MTM}. This is referred to, in this work, as \gls{TransM+MTM}. The comparable results with the rest of the reviewed approaches are depicted in Table \ref{tab:cifar-10_ssdl}.

An alternative approach for the consistency function  
was implemented in \cite{tran2019semi}, with a model named by the
authors as \gls{SESEMI}. The consistency function is fed by what
the authors defined as a self-supervised branch. This branch aims to learn simple
image transformations or pretext tasks, such as image rotation. The authors claim that their
model is easier to calibrate than the \gls{MTM} , by just using an unsupervised signal weight of $\lambda=1$. The intersected results of \gls{SESEMI} with the rest of the reviewed methods are detailed in Table  \ref{tab:cifar-10_ssdl}.

In \cite{berthelot2019mixmatch}, the authors proposed a novel  \gls{SSDL} method known as MixMatch. This method implements a consistency loss which first calculates a soft pseudo-label for each unlabelled observation. Those soft pseudo-labels are the result of averaging the model response to a number of transformations of the input $\textbf{x}_{j}$ $\widehat{\textbf{y}}{}_{j}=\frac{1}{\mathcal{T}}\sum_{\eta=1}^{\mathcal{T}}f_{\textbf{w}}\left(\Psi^{\eta}\left(\textbf{x}_{j}\right)\right)$. In such equation, $\mathcal{T}$ refers to the number of transformations of the input image (i.e., image rotation, cropping, etc.). The specific image transformation is represented in $\Psi^{\eta}$.  The authors in \cite{berthelot2019mixmatch} recommend to use  $\mathcal{T} = 2$.  Later, the obtained soft pseudo-label is sharpened, in order to decrease its entropy and under-confidence of the pseudo-label. For this, a parameter $\rho$ is used within the softmax of the output $\widehat{\textbf{y}}{}_{j}$: $s\left(\widehat{\textbf{y}},\rho\right)_{i}=\frac{\widehat{y}_{i}^{1/\rho}}{\sum_{j}\widehat{y}_{j}^{1/\rho}}$.  The dataset $\widetilde{S}_{u}=\left(X_{u},\widetilde{Y}\right)$ contains the sharpened soft pseudo-labels, where $\widetilde{Y}=\left\{ \widetilde{\textbf{y}}_{1},\widetilde{\textbf{y}}_{2},\ldots,\widetilde{\textbf{y}}_{n_{u}}\right\}$. The authors of MixMatch found that data augmentation is very important to improve its performance. Taking this into account, the authors implemented the MixUp methodology to augment both the labelled and unlabelled datasets \cite{zhang2017mixup}. This is represented as follows:  $\left(S'_{l},\widetilde{S}'_{u}\right)=\Psi_{\textrm{MixUp}}\left(S_{l},\widetilde{S}_{u},\alpha\right)$.  The MixUp generates new observations through a linear interpolation between different combinations of both the labelled and unlabelled data. The labels for the new observations are also lineally interpolated, using both the labels and the pseudo-labels (for the unlabelled data).

Mathematically, MixUp takes  two pseudo-labelled or  labelled data pairs $\left(\textbf{x}_{a},y_{a}\right)$ and $\left(\textbf{x}_{b},y_{b}\right)$, and generates the augmented datasets $ \left(S'_{l},\widetilde{S}'_{u}\right)$. These augmented datasets are used by MixMatch, to train  neural network with parameters $\textbf{w}$  through the minimization of the following loss function:

\[
\mathcal{L}\left(S,\textbf{w}\right)=\sum_{\left(\textbf{x}_{i},\textbf{y}_{i}\right)\in S'_{l}}\mathcal{L}_{l}\left(\textbf{w},\textbf{x}_{i},\textbf{y}_{i}\right)+
\]
\begin{equation}
\gamma r(t) \sum_{\left(\textbf{x}_{j},\widetilde{\textbf{y}}_{j}\right)\in\widetilde{S}'_{u}}\mathcal{L}_{u}\left(\textbf{w},\textbf{x}_{j},\widetilde{\textbf{y}}_{j}\right)
\end{equation}

The labelled loss term $\mathcal{L}_l$, can be implemented with a cross-entropy function, as recommended in \cite{berthelot2019mixmatch}; $\mathcal{L}_{l}\left(\textbf{w},\textbf{x}_{i},\textbf{y}_{i}\right)=H_{\textrm{CE}}\left(\textbf{y}_{i},f_{\textbf{w}}\left(\textbf{x}_{i}\right)\right)$. Regarding the unlabelled loss term, an  Euclidean distance was tested by the authors in \cite{berthelot2019mixmatch}  $\mathcal{L}_{u}\left(\textbf{w},\textbf{x}_{j},\widetilde{\textbf{y}}_{j}\right)=\left\Vert \widetilde{\textbf{y}}_{j}-f_{\textbf{w}}\left(\textbf{x}_{j}\right)\right\Vert$.  In the MixMatch loss function, the coefficient $r(t)$ is implemented  as a ramp-up function which augments the weight of the unlabelled loss term as  $t$ increases.  The parameter $\gamma$ controls the overall influence of the unlabelled loss term. In Table  \ref{tab:cifar-10_ssdl}, the results yielded in \cite{berthelot2019mixmatch} are depicted for the \ \gls{CIFAR-10} dataset.

In \cite{berthelot2019remixmatch} an extension of the MixMatch algorithm was developed, referred to as ReMixMatch. Two main modifications were  proposed: a distribution alignment procedure and a more extensive use of data augmentation. Distribution alignment consists of the normalization of each prediction using both the  running average prediction of each class (in a set of previous model epochs) and the marginal label distribution using the labelled dataset. This way, soft pseudo-label estimation accounts for the label distribution and previous label predictions,  enforcing soft pseudo-label consistency with both distributions.   The extension of the previous simple data-augmentation step implemented in the original MixMatch algorithm (where flips and crops were used) consists of two methods. They are referred to as anchor augmentation and CTAugment by the authors. The empirical evidence gathered by the authors when implementing stronger data augmenting transformations (i.e. gamma and brightness modifications, etc.) in MixMatch showed a performance deterioration. This is caused by the larger variation in the model output for each type of strong transformation, making the pseudo-label less meaningful. To circumvent this, the authors proposed an \textit{augmentation anchoring} approach. It uses the same pseudo-labels estimated when using a weak transformation, for the $\mathcal{T}'$ strong transformations. Such strong transformations are calculated through a modification of the auto-augment algorithm. Auto-augment originally uses reinforcement learning to find the best resulting augmentation policy (set of transformations used) for the specific target problem \cite{cubuk2018autoaugment}. To simplify its implementation for small labelled datasets, the authors in \cite{berthelot2019remixmatch} proposed a  modification referred to as CTAugment. It estimates the likelihood of generating a correctly classified image, in order to generate images that are unlikely to result in wrong predictions. The performance reported in the executed benchmarks of ReMixMatch, showed an accuracy gain ranging from 1\% to 6\%, when compared to the original MixMatch algorithm. No statistical significance tests were reported. Comparable results to other methods reviewed in this work for the \gls{CIFAR-10} dataset are shown in Table \ref{tab:cifar-10_ssdl}. 

More recently, an \gls{SSDL} method referred to as FixMatch was proposed in \cite{sohn2020fixmatch}. The authors argue the proposition of a simplified \gls{SSDL} method compared to other techniques.  FixMatch is based upon pseudo-labelling and consistency regularized \gls{SSDL}. The loss function uses a cross-entropy labelled loss term, along with weak augmentations for the labelled data. For the unlabelled loss term, the cross-entropy is also used, but for the \textit{strongly} unlabelled observations with its corresponding pseudo-label. The soft pseudo-label is calculated using weak transformations, taking the maximum logit of the model output. Therefore no model output sharpening is done, unlike MixMatch. Strong augmentations are tested using both \gls{RA} \cite{cubuk2020randaugment} and \gls{CTA} \cite{berthelot2019remixmatch}. For benchmarking FixMatch, the authors used the \gls{CIFAR-10} (40, 250, 4000 labels), \gls{CIFAR-100} (400, 2500 and 10000 labels), \gls{SVHN} (40, 250, 1000 labels)  and \gls{STL-10} (1000 labels) datasets. For all the methods, variations of the Wide-ResNet \gls{CNN} backbone were used.  The average accuracy for each test configuration was similar to the results yielded  by ReMixMatch, with no statistical significance tests performed. Comparable results yielded by FixMatch are depicted in Table \ref{tab:cifar-10_ssdl}.

\subsubsection{Adversarial augmentation based regularization}\label{subsec:Augmented-regularized-loss}

Recent advances in deep generative networks for learning data distribution
has encouraged its usage in \gls{SSDL}  architectures \cite{goodfellow2014generative}.
Regularized techniques usually employ basic data augmentation pipelines,
in order to evaluate the consistency term $\mathcal{L}_{u}$. However,
generative neural networks can be used to learn the distribution of
labelled and unlabelled data and generate entirely new observations. These are
categorized as \gls{GaNC-SSDL}. Learning a good approximation of
data distribution $\mathbf{Pr}_{\textbf{x}\sim\mathcal{C}}\left(\textbf{x}\right)$
allows the artificial generation of new observations. The observations can be added
to the unlabelled dataset $S^{(u)}$, or the very same adversarial training might lead to a refined set of model parameters. 

In \cite{springenberg2015unsupervised}, the generative network architecture
 was extended for \gls{SSDL}, by implementing a discriminator
function $f_{\textbf{w}_{d}}^{\left(d\right)}$ able to estimate not only if an observation
$\textbf{x}_{i}$ belongs or not to one of the  classes to discriminate from,
but also to which specific class it belongs to. The model was named
by the authors as \gls{CAT-GAN}, given the capacity of the discriminator
to perform ordinary $1-K$ classification.  Therefore, $f_{\textbf{w}_{d}}^{\left(d\right)}$ is able to estimate the density function of an unlabelled observation $\textbf{x}_{i}\in S^{(u)}$, $\hat{\textbf{y}}_{i}=f_{\textbf{w}_{d}}^{\left(d\right)}\left(\textbf{x}_{i}\right)$. The discriminator model implements a semi-supervised loss function: \begin{equation} \mathcal{L}^{\left(d\right)}\left(f_{\textbf{w}_{d}}^{\left(d\right)},\textbf{x}_{i}\right)=\mathcal{L}_{l}^{\left(d\right)}\left(f_{\textbf{w}_{d}}^{\left(d\right)},\textbf{x}_{i},\textbf{y}_{i}\right)+\mathcal{L}_{u}^{\left(d\right)}\left(f_{\textbf{w}_{d}}^{\left(d\right)},\textbf{x}_{i}\right) \label{eq:CAT_d} \end{equation} with $\mathcal{L}_{l}^{\left(d\right)}\left(\textbf{w}_{d},\textbf{x}_{i}\right)=H_{\textrm{CE}}\left(\textbf{y}_{i},f_{\textbf{w}_{d}}^{\left(d\right)}\left(\textbf{x}_{i}\right)\right),$ where $H_{\textrm{CE}}$ is the cross entropy. The unsupervised discriminator term $\mathcal{L}_{u}^{\left(d\right)}$ was designed for maximizing the certainty for unlabelled observations and minimizing it for artificial observations. The authors also included a term for encouraging imbalance correction for the $K$ classes. The proposed method in \cite{springenberg2015unsupervised} was tested using the  \gls{CIFAR-10} and \gls{MNIST} datasets for \gls{SSDL}. It was compared only against the \gls{Pi-M}, with marginally better average results and no statistical significance analysis of the results. However, the \gls{CAT-GAN} served as a foundation for posterior work on using generative deep models for \gls{SSDL}.

A breakthrough improvement in training $f_{\textbf{w}_{d}}^{\left(d\right)}$ and $f_{\textbf{w}_{g}}^{\left(g\right)}$ models was achieved in \cite{salimans2016improved}, aiming to overcome the difficulty of training complementary loss functions with a stochastic gradient descent algorithm. This problem is known as the Nash equilibrium dilemma. The authors yielded such improvement, through a feature matching loss for the generator $f_{\textbf{w}_{g}}^{\left(g\right)}$, which seeks to make the generated observations match the statistical moments of a real sample from training data $S$. The enhanced trainability of the \gls{FM-GAN} was tested in a semi-supervised learning setting. The semi-supervised loss function implements an unsupervised term $\mathcal{L}_{u}^{\left(d\right)}\left(f_{\textbf{w}_{d}}^{\left(d\right)},\textbf{x}_{i}\right)$ which aims to maximize the discriminator success rate in discriminating both unlabelled real observations $\textbf{x}_{i}\in S^{(u)}$ and artificially generated ones. The discriminator model $f_{\textbf{w}_{d}}^{\left(d\right)}\left(\textbf{x}_{i}\right)$ outputs the probability of the observation $\textbf{x}_{i}$ belonging to one of the real   classes. Also in this work, the authors showed how achieving a good semi-supervised learning accuracy (thus a good discriminator), often yields a poor generative performance. Authors suggested that a bad generator describes better \gls{OOD} data, improving the overall model robustness. Intersected results with the rest of the reviewed \gls{SSDL} methods are depicted in Table \ref{tab:cifar-10_ssdl}.

In \cite{dai2017good}, the authors further explored the inverse relationship
between generator and semi-supervised discriminate performance with
the \gls{Bad-GAN}. The experiments   showed
how a \emph{bad} generator $f_{\textbf{w}_{g}}^{\left(g\right)}$, created observations
out of the distribution of the concept class $\mathbf{Pr}_{\textbf{x}\sim\mathcal{C}}$
enhancing the performance of the discriminator $f_{\textbf{w}_{d}}^{\left(d\right)}$
for semi-supervised learning. More specifically, the generator loss
$\mathcal{L}^{\left(g\right)}$ is encouraged to maximize the Kullback-Leibler
distance to the sampled data distribution. This enforces the boundaries built by the discriminator $f_{\textbf{w}_{d}}^{\left(d\right)}$ for distractor observations.  In \cite{li2019semi}
a comparison between the triple generative network proposed in \cite{chongxuan2017triple}
and the bad generator \cite{dai2017good} was done. No conclusive results
were reached, leading the authors to suggest a combination of the approaches
to leverage accuracy.  For comparison purposes with  related work, results with the \gls{CIFAR-10}  are described in Table \ref{tab:cifar-10_ssdl}.

Later, in \cite{qiao2018deep}, the authors proposed a co-training adversarial
regularization approach for the \gls{Co-GAN}, making use of the consistency
assumption of two different models $f_{\textbf{w}_{d_1}}^{\left(d_1\right)}$ and $f_{\textbf{w}_{d_2}}^{\left(d_2\right)}$. Each model is trained with a different view from the same observation
$\textbf{x}_{i}=\left\langle \textbf{x}_{i}^{\left(v_{1}\right)},\textbf{x}_{i}^{\left(v_{2}\right)}\right\rangle $.
A general loss function $\mathcal{L}\left(S\right)=\mathcal{L}_{l}\left(S_{l}\right)+\mathcal{L}_{u}\left(S^{(u)}\right)$
is minimized, with the unsupervised loss function defined as 
\begin{equation}
\mathcal{L}_{u}\left(f_{\textbf{w}},S^{(u)}\right)=\lambda_{\textrm{cot}}\mathcal{L}_{\textrm{cot}}\left(f_{\textbf{w}},S^{(u)}\right)+\lambda_{\textrm{dif}}\mathcal{L}_{\textrm{dif}}\left(f_{\textbf{w}},\textbf{z}\right).\label{eq:Co-GANN}
\end{equation}
The term $\mathcal{L}_{\textrm{cot}}$ measures
the expected consistency of the two models using two views from the
same observation, through the Jensen-Shannon divergence. In $\mathcal{L}_{\textrm{dif}}\left(\textbf{z}\right)$,
each model artificially generates observations to deceive the other
one. This stimulates a view difference between the models, to avoid them
collapsing into each other. The coefficients $\lambda_\textrm{cot}$ and $\lambda_\textrm{dif}$ weigh  the contribution of each term.  Therefore, for each view, a generator
is trained and the models $f_{\textbf{w}_{d_1}}^{\left(d_1\right)}$ and $f_{\textbf{w}_{d_2}}^{\left(d_2\right)}$
play the detective role. The proposed method out performed \gls{MTM},
\gls{TEM} and the \gls{Bad-GAN} according to \cite{qiao2018deep}. Experiments
were performed with more than two observation views, generalizing
the model for a multi-view layout $\textbf{x}_{i}=\left\langle \textbf{x}_{i}^{\left(v_{1}\right)},\textbf{x}_{i}^{\left(v_{2}\right)}\right\rangle $. The best performing model implemented 8 views, henceforth referred in this document as \gls{Co-8-GAN}. We include results of the benchmarks done in \cite{qiao2018deep} with the \gls{SVHN} and \gls{CIFAR-10} datasets. The authors did not report any statistical significance analysis of the provided results.

The \gls{Triple-GAN} \cite{chongxuan2017triple} addressed the aforementioned inverse relationship between generative and semi-supervised classification performance, by training three different models, detailed as follows. First, a classifier $f_{\textbf{w}_{c}}^{\left(c\right)}\left(\textbf{x}_{i}\right)$ which learns the data distribution $\mathbf{Pr}_{\textbf{x}\sim\mathcal{C}}$ and outputs pseudo-labels for artificially generated observations. Secondly, a class-conditional generator $f_{\textbf{w}_{g}}^{\left(g\right)}$ able to generate observations for each individual class. Thirdly, a discriminator $f_{\textbf{w}_{d}}^{\left(d\right)}\left(\textbf{x}_{i}\right)$, which rejects observations out of the labelled classes. The architecture uses pseudo-labelling, since the discriminator uses the pseudo labels of the classifier $\hat{\textbf{y}}_{i}=f_{\textbf{w}_{c}}^{\left(c\right)}\left(\textbf{x}_{i}\right)$. Nevertheless, a consistency regularization was implemented in the classifier loss $\mathcal{L}^{\left(c\right)}$. The results using  \gls{CIFAR-10} with the settings also tested in the rest of the reviewed works are depicted in tables \ref{tab:cifar-10_ssdl}.

\subsubsection{Graph based regularization}\label{subsec:Graph-based-regularized}

\gls{GR-SSDL} is based on previous graph based regularization techniques \cite{goldberg2007dissimilarity}. The core idea of \gls{GR-SSDL} is to preserve mutual distance from observations in
the dataset $S$ (for both labelled and unlabelled) in a new feature
space. An embedding is built through a mapping function $\check{\textbf{x}}_{i}=h_{\textbf{w}_{\textrm{FE}}}^{\left(\textrm{FE}\right)}\left(\textbf{x}_{i}\right)$
which reduces the input dimensionality $d$ to $\check{d}$. The  mutual
distance of the observations in the original input space $\textbf{x}_{i}\in\mathbb{R}^{d}$
, represented in the matrix $W\in\mathbb{R}^{n\times n}$, with
$W_{i,j}=\delta\left(\textbf{x}_{i},\textbf{x}_{j}\right)\label{eq:Woriginal}$
is meant  to be preserved in the new feature space $\check{\textbf{x}}_{i}\in\mathbb{R}^{\check{d}}$. The multidimensional
scaling algorithm  developed in \cite{kruskal1964multidimensional} is one of the  first approaches to preserve the mutual distance of the embeddings of the observations.

More recently, in \cite{luo2018smooth}, a graph-based regularization
was implemented in the \gls{SNTGM}. The model aims to smooth the consistency
of the classifier along the observations in a cluster, and not only the artificially created observations by the previous consistency-based regularization techniques. The proposed approach implements both a consistency based regularization $\mathcal{L}_{c}$ with weight $\lambda_1$, and a guided embedding $\mathcal{L}_{e}$ with coefficient $\lambda_2$:
\begin{equation}
\mathcal{L}_{u}\left(f_{\textbf{w}},\textbf{x}_{i},\textbf{x}_{j},W_{i,j}\right)=\lambda_{1}\mathcal{L}_{c}\left(f_{\textbf{w}},\textbf{x}_{i}\right)+\lambda_{2}\mathcal{L}_{e}\left(f_{\textbf{w}},\textbf{x}_{i},\textbf{x}_{j},W\right)\label{eq:GraphRegu-1}
\end{equation}
where $\textbf{x}_{i},\textbf{x}_{j} \in S^{(u)}$.  $\mathcal{L}_{c}$ measures the prediction consistency, by using previous approaches in consistency based regularized techniques. The term $\mathcal{L}_{e}$ implements
the observation embedding, with a $\gamma$ margin-restricted distance.
To build the neighbourhood matrix $W$, the authors in \cite{luo2018smooth}  used  label information instead of computing the distance between the observations.  Regarding
unlabelled observations in $S^{(u)}$, the authors estimated the output of the teacher to be 
$\hat{y}_{i}=f_{\textbf{w}'}\left(\Psi^{\eta}\left(\textbf{x}_{i}\right)\right)$.
Thus, the neighbourhood matrix is given as follows:
\begin{equation}
W_{i,j}=\begin{cases}
1 & \textrm{if }\hat{y}_{i}=\hat{y}_{j}\\
0 & \textrm{if }\hat{y}_{i}\neq\hat{y}_{j}
\end{cases}.\label{eq:GraphW}
\end{equation}

The loss term $\mathcal{L}_{e}$ encourages similar representations for observations within the same class and higher difference for representations of different classes. The algorithm was combined and tested with a \gls{Pi-M} and
\gls{VATM} consistency functions, henceforth \gls{SNTGM+Pi-M} and
\gls{SNTGM+VATM}, respectively.

\begin{table*}[]
\center
\resizebox{8cm}{!}{
\begin{tabular}{c|c|ccc}
\textbf{Model}       & \textbf{Category}                & \textbf{$n_l=2000$}                                          & \textbf{$n_l=4000$}                                                  & \textbf{$n_l=5000$}                                        \\ \hline
Supervised only      & Supervised                       & 33.94$\pm$0.73\cite{tarvainen2017mean} & 20.02$\pm$0.6\cite{tarvainen2017mean}         &  18.02$\pm$0.6\cite{tarvainen2017mean}                                                      \\ \hline
\gls{Pi-M}          &                                  & 18.02$\pm$0.6\cite{tarvainen2017mean}    & 13.2$\pm$0.27\cite{laine2016temporal}              & 6.06$\pm$0.11\cite{laine2016temporal}                                                         \\
\gls{TEM}           &                                  & -     & 12.16$\pm$0.24\cite{laine2016temporal} & 5.6$\pm$0.1 \cite{laine2016temporal}                                                          \\

\gls{VATM+EM}       &                                  & -                                                           & 13.15$\pm$0.21\cite{miyato2018virtual}           & -                                                          \\
\gls{VATM}          &                                  & -                                                           & 14.87$\pm$0.13\cite{miyato2018virtual}            & -                                                          \\
\gls{MTM}           &                                  & 15.73$\pm$0.31\cite{tarvainen2017mean}     & 12.31$\pm$0.28\cite{tarvainen2017mean} & 5.94$\pm$0.15\cite{qiao2018deep,tarvainen2017mean}                                                          \\
\gls{SESEMI}        &                                  & 14.22$\pm$0.27\cite{tran2019semi}           & 11.65$\pm$0.13\cite{tran2019semi}                 & -                                                          \\
\gls{METM}          &                                  & -  & 11.29$\pm$0.24\cite{sajjadi2016regularization}            & - \\
\gls{TransM}        &  \gls{RC-SSDL}   & 14.65$\pm$0.33\cite{shi2018transductive} & 10.9$\pm$0.23\cite{shi2018transductive} & 5.2$\pm$0.14\cite{shi2018transductive} \\
\gls{TransM+MTM}    &                                  & 13.54$\pm$0.32\cite{shi2018transductive} & 9.3$\pm$0.55\cite{shi2018transductive} & 5.19$\pm$0.14\cite{shi2018transductive} \\
\gls{MeM}           &                                  & -                                                           & 11.91$\pm$0.22\cite{chen2018semi}                   & -                                                          \\
MixMatch             &                                  & -                                                           & 6.42$\pm$0.10\cite{berthelot2019mixmatch}                                                         & -                                                          \\
ReMixMatch           &                                  & -                                                           & 4.72$\pm$0.13\cite{berthelot2019remixmatch}                                                      & -                                                          \\
FixMatch(\gls{RA})  &                                  & -                                                           & 4.31$\pm$0.15                                                        & -                                                          \\
FixMatch(\gls{CTA}) &                                  & -                                                           & 4.26$\pm$0.05                                                        & -                                                          \\ \hline
\gls{SNTGM+VATM}    &  \gls{GR-SSDL}   & -                                                           & 12.49$\pm$0.36\cite{luo2018smooth}                 & -                                                          \\
\gls{SNTGM+Pi-M}      &                                  & -                                                           & 13.62$\pm$0.17\cite{luo2018smooth}                 & -                                                          \\ \hline
\gls{FM-GAN}        &                                  & -  & 18.63$\pm$2.32\cite{cicek2018saas}            & - \\
\gls{CAT-GAN}      &                                  & -                                                           & 19.58$\pm$0.58\cite{springenberg2015unsupervised}                      & -                                                          \\
\gls{Co-8-GAN}      &  \gls{GaNC-SSDL} & -                                                           & 8.35$\pm$0.06\cite{qiao2018deep}                    & -                                                        \\
\gls{Bad-GAN}       &                                  & -            & 14.41$\pm$0.3\cite{dai2017good}                   & -                                                          \\
\gls{Triple-GAN}    &                                  & -            & 16.99$\pm$0.36\cite{chongxuan2017triple}                   & -                                                          \\ \hline
\gls{Tri-Net}       &                                  & -                                                           & 8.45$\pm$0.22\cite{dong2018tri}                   & -                                                          \\
\gls{SaaSM}         &  \gls{PLT-SSDL}   & -                                                           & 10.94$\pm$0.07\cite{cicek2018saas}                 & -                                                          \\
\gls{TriNet+Pi}     &                                  & -                                                           & 8.3$\pm$0.15\cite{dong2018tri}                     & -                                                         
\end{tabular}}\caption{\gls{SSDL} error rates (the lower the better) from literature of state of the art methods, using the \gls{CIFAR-10} dataset. As number of labels, $n_l=2000$, $n_l=4000$ and $n_l=5000$ were the most frequently used in the literature. }\label{tab:cifar-10_ssdl}
\end{table*}

\section{Dealing with distribution mismatch in SSDL}\label{subsec:ssdl_mismatch_state_of_art}
In \cite{oliver2018realistic} and \cite{calderon2020mixmood}, extensive evaluation of the distribution mismatch setting is developed. The authors agreed upon its decisive impact in the performance of \gls{SSDL} methods and the consequent importance of increase their robustness to such phenomena. 
\gls{SSDL} methods designed to deal with the distribution mismatch between $S^{(u)}$ and $S^{(l)}$ often use ideas and concepts from \gls{OOD} detection techniques. Most methods for  \gls{SSDL} that are  robust to distribution mismatch calculate a weight or a coefficient referred to as the function $\mathcal{H}\left(\textbf{x}_{j}^{u}\right)$ in this article, to score how likely the unlabelled observation $\textbf{x}_{j}^{u}$ is \gls{OOD}. The score can be used to either completely discard $\textbf{x}_{j}^{u}$ from the unlabelled training dataset (referred to as hard thresholding in this work) or to weigh it (soft thresholding).  Thresholding the unlabelled dataset can take place as a data pre-processing step or in an online fashion during training.

Therefore, we first review modern approaches for \gls{OOD} detection using deep learning in \Cref{subsec:OOD_detection}. Later we  address state of the art \gls{SSDL} methods that are robust to distribution mismatch in \Cref{subsec:SSDL_robust}.

\subsection{\gls{OOD} Data Detection}\label{subsec:OOD_detection}

\gls{OOD} data detection is a classic challenge faced in machine learning applications. It corresponds to the detection of data observations which are far from the training dataset distribution \cite{hendrycks2016baseline}. Individual and collective outlier detection are particular problems of \gls{OOD} detection  \cite{singh2012outlier}. Other particular \gls{OOD} detection settings have been tackled in the literature such as    novel data and anomaly detection \cite{perera2019deep}  and  infrequent event detection \cite{hamaguchi2019rare,amodei2016concrete}. Well studied and known concepts have been developed within the pattern recognition community related to \gls{OOD} detection. Some of them are kernel representations \cite{10.1023/B:MACH.0000008084.60811.49},   density estimation \cite{markou2003novelty}, robust moment estimation \cite{doi:10.1080/01621459.1984.10477105} and  prototyping \cite{markou2003novelty}.

The more recent developments in the burgeoning field of deep learning for image analysis tasks have boosted the interest in developing \gls{OOD} detection methods for deep learning architectures. According to our literature survey, we found that \gls{OOD} detection methods for deep learning architectures can be classified into the following categories: \gls{DNN} output based and \gls{DNN} feature space based. In the next subsections we proceed to describe the most popular methods within each category.

\subsubsection{DNN output based}

In \cite{DBLP:journals/corr/HendrycksG16c} the authors proposed a simple method known as \gls{ODIN} to score input observations according to its \gls{OOD} probability. The proposed method by the authors implements a confidence score based upon the \gls{DNN} model's output which is transformed using a softmax layer. The maximum softmax value of all the units is associated with the model's confidence. The authors argued that this scores is able to distinguish in-distribution from \gls{OOD} data.

 More recently, in \cite{liang2018enhancing}, the authors argued that using the softmax output of a \gls{DNN} model to estimate \gls{OOD} probability can be often a misleading measure in non calibrated models. Therefore, the authors in \cite{liang2018enhancing} proposed a \gls{DNN} calibration method. This method implements a temperature coefficient which aims to improve the model's output discriminatory power between  \gls{OOD} and \gls{IOD} data. In \cite{liang2018enhancing} the authors tested \gls{ODIN} against the softmax based \gls{OOD} score proposed in \cite{DBLP:journals/corr/HendrycksG16c}, with significantly better results obtained by \gls{ODIN}. 
 
 An alternative approach for \gls{OOD} detection using the \gls{DNN}'s output is the popular approach known as \gls{MCD} \cite{loquercio2020general,kendall2017uncertainties}. This approach uses the distribution of $N$ model forward passes (input evaluation), using the same input observation with mild transformations (noise, flips, etc.) or injecting noise to the model using a parameter drop-out. The output distribution is used to calculate distribution moments (variance usually) or other scalar distribution descriptors such as the entropy. This idea has been implemented in \gls{OOD} detection settings, as \gls{OOD} observations might score higher entropy or variance values \cite{jin2019augmenting,sedlmeier2020uncertainty}.

 \subsubsection{DNN's feature space based}
 
 Recently, as an alternative approach for \gls{OOD} detection, different methods use the feature or latent space for \gls{OOD} detection. In \cite{lee2018simple} the authors propose the usage of the Mahalanobis distance in the feature space between the training dataset and the input observation. Therefore the covariance matrix and a mean observation is calculated from the training data (in-distribution data). By using the Mahalanobis distance, the proposed method by the authors assume a Gaussian distribution of the training data. Also, the proposed method was tested mixed with the \gls{ODIN} calibration method previously discussed.  The authors reported a superior performance of their method over the softmax based score proposed in \cite{DBLP:journals/corr/HendrycksG16c} and \gls{ODIN} \cite{liang2018enhancing,DBLP:journals/corr/HendrycksG16c}. However no statistical significance analysis of the results was carried out.

 In \cite{vanuncertainty} another feature space based was proposed, referred to as \gls{DUQ} by the authors. The  proposed method was tested for both uncertainty estimation and \gls{OOD} detection. It consists in calculating a centroid for each one of the classes within the training dataset (\gls{IOD} dataset). Later, for each new observation where either uncertainty estimation or \gls{OOD} detection is intended to be used, the method calculates  the distance to each centroid. The shortest distance is used as either uncertainty or \gls{OOD} score.  \gls{DUQ} performance for \gls{OOD} detection was compared against a variation of the \gls{MCD} approach, with a an ensemble of networks for \gls{OOD} detection. The authors claimed a better performance of \gls{DUQ} for \gls{OOD} detection, however no statistical analysis of the results was done.  The benchmark consisted in using  \gls{CIFAR-10} as an \gls{IOD} dataset and \gls{SVHN} as a \gls{OOD} dataset. Therefore, as usual in \gls{OOD} detection benchmarks, the  \textit{unseen classes} setting for the \gls{IID} assumption violation was tested.

\subsection{Semi-supervised Deep Learning methods robust to distribution mismatch}\label{subsec:SSDL_robust}

In the literature, there are two most commonly studied causes for the violation of the \gls{IID}  assumption. The first one is the  prior probability shift (different distribution of the labels) between $S^{(u)}$ and $S^{(l)}$. Novel methods proposed to deal with this challenge are described in this section. The other cause for the \gls{IID} violation assumption is the \textit{unseen class setting}, which has been more widely studied. State of the art methods are also discussed in this section.

 \subsubsection{Unseen classes as a cause for the distribution mismatch}\label{subsec:unseen}
 Most of the \gls{SSDL} methods designed to deal with distribution mismatch have been tested using a labelled dataset with different classes (usually less) from the unlabelled dataset. For example, in this setting, for $S^{(l)}$ the \gls{SVHN} is used, and for $S^{(u)}$  a percentage of the sample is drawn from the \gls{CIFAR-10} dataset, and the rest from the \gls{SVHN} dataset. In this context, the dataset \gls{CIFAR-10} is often referred to as the \gls{OOD} data contamination source. Benchmarks with varying degrees of data contamination for \gls{SSDL} with distribution mismatch can be found in literature.  In this section we describe the most recent approaches for \gls{SSDL} under distribution mismatch with unseen classes in the literature.
 
 In \cite{nair2019realmix} an \gls{SSDL} method for dealing with distribution mismatch was developed. The authors refer to this method as RealMix.  It was proposed as an extension of the MixMatch \gls{SSDL} method. Therefore, it uses the consistency based regularization with augmented observations, and the MixUp data augmentation method implemented in MixMatch. For distribution mismatch robustness, RealMix uses the softmax of the output from the model as a confidence value, to score each unlabelled observation. During training, in the loss function, the unlabelled observations are masked out using such confidence score. The $\phi$ percent of  unlabelled observations with the lowest confidence scores are discarded at each training epoch.  To test their method, the authors deployed a benchmark based upon \gls{CIFAR-10} with a disjoint  set of classes for  $S^{(l)}$ and $S^{(u)}$. The reported results showed a slight accuracy gain of the proposed method against other \gls{SSDL} approaches not designed for distribution mismatch robustness. A fixed number of labelled observations and \gls{CNN} backbones were used.  No statistical significance tests over the results were done. RealMix can be categorized as a \gls{DNN} output based \gls{OOD} scoring method. The thresholding is done during training, several times, using binary or hard thresholding (keep or discard). The testing can be considered limited as the \gls{OOD} contamination source causes a hard distribution mismatch.
 
 More recently, the method known as \gls{UASD}, was proposed in \cite{chen2020semi} for \gls{SSDL} distribution mismatch robustness. \gls{UASD} uses an unsupervised regularized loss function. For each unlabelled observation, a pseudo-label is estimated as the average label from an ensemble of models. The ensemble is composed of past models yielded in previous training epochs.   Similar to RealMix, \gls{UASD} uses the output of a \gls{DNN} model to score each unlabelled observation. However, to increase the robustness of such confidence score, \gls{UASD} uses the ensemble of predictions from past models, to estimate the model's confidence over its prediction for each unlabelled observation. The maximum logits of the ensemble prediction is used as the confidence score. Therefore we can categorize the \gls{UASD} method as a \gls{DNN} output based approach.  Also in a similar fashion to RealMix, the estimated scores are used for hard-thresholding the unlabelled observations.  In a resembling trend  to RealMix, the authors of the \gls{UASD} method evaluated their approach using the \gls{CIFAR-10} dataset.  $S^{(l)}$ includes 6 classes of animals, and $S^{(u)}$ samples other  4 classes from \gls{CIFAR-10}, with a varying degree of class distribution mismatch. Only five runs were performed to approximate the error-rate distribution, and no statistical analysis was done for the results. No varying number of labelled observations, or different \gls{DNN} backbones were tested.  \gls{UASD} was compared with \gls{SSDL} methods not designed for distribution mismatch robustness. From the reported results, an accuracy gain of up to 6 percent over previous \gls{SSDL} methods was yielded by \gls{UASD}, when facing heavy distribution mismatch settings.
 
 In \cite{chen2020semi}, an \gls{SSDL} approach to deal with distribution mismatch was introduced. The authors refer to their proposed approach as \gls{D3SL}. It implements an unsupervised regularization, through the mean square loss between the prediction of unlabelled observation and its noisy modification. An  observation-wise weight  for each unlabelled observation is implemented, similar to RealMix and \gls{UASD}. However, the weights for the entire unlabelled dataset are calculated using an error gradient optimization approach.  Both the model's parameters and the observation-wise weights are estimated in two nested optimization steps. Therefore, we can categorize this method as a gradient optimized scoring of the unlabelled observations. The weights are continuous or non-binary values, therefore we can refer to this method as a softly-thresholded one. According to the authors, this increases training time up to $3\times$. The testing benchmark uses the \gls{CIFAR-10} and \gls{MNIST} datasets. For both of them, 6 classes are used to sample $S^{(l)}$, and the remaining for $S^{(u)}$. Only a Wide ResNet-28-10 \gls{CNN} backbone was used with a fixed number of labels. A varying degree of \gls{OOD} contamination was tested. The proposed \gls{D3SL} method was compared with generic \gls{SSDL} methods, therefore ignoring previous \gls{SSDL} robust methods to distribution mismatch. From the reported results, an averaged accuracy gain of around 2\% was yielded by the proposed method under the heaviest \gls{OOD} data contamination settings, with no statistical significance reported. Only five runs were done to report such averaged error-rates. 
 
 A similar gradient optimization based method to \gls{D3SL}  can be found in \cite{yu2020multi}. The proposed method is referred to as a  \gls{MTCF} by the authors. Similar to previous methods, \gls{MTCF} defines an \gls{OOD} score for the unlabelled observations, as an extension to the MixMatch algorithm \cite{berthelot2019mixmatch}. Such scores are alternately optimized together with the \gls{DNN} parameters, as seen in  \gls{D3SL}. However, the optimization problem is perhaps more simple than the  \gls{D3SL}, as the \gls{OOD} scores are not optimized in a gradient descent fashion directly. Instead, the \gls{DNN} output is used as the \gls{OOD} score. The usage of a loss function that includes the \gls{OOD} scores, enforces a new \textit{condition} to the optimization of the \gls{DNN} parameters. This   is referred to  as a curriculum multi-task learning framework by the authors in \cite{yu2020multi}. The proposed method was tested in what the authors defined as an \gls{Open-Set-SSLS}, where different \gls{OOD} data contamination sources were used. Regarding the specific benchmarking settings, the authors only tested a Wide ResNet \gls{DNN} backbone, to compare a baseline MixMatch method to their proposed approach. No comparison with other \gls{SSDL} methods was performed. The authors used two  \gls{IOD} datasets: \gls{CIFAR-10} and \gls{SVHN}. Four different \gls{OOD} datasets were used: Uniform,  Gaussian noise, \gls{TIN} and \gls{LSUN}. The average of the last 10 checkpoints of the model training, using the same partitions was reported (no different partitions were tested). A fixed \gls{OOD} data contamination degree was tested. The reported accuracy gains went from 1\% to 10\%. The usage of the same data partitions inhibited an appropriate statistical analysis of the results. 
 
 In the same trend, authors in \cite{zhao2020robust} proposed a gradient optimization based method to calculate the observation-wise weights for data in $S^{(u)}$. Two different gradient approximation methods were tested: \gls{IF} and \gls{MetaA}. Authors argue that finding the weights for each unlabelled observation in a large sample $S^{(u)}$ is an intractable problem. Therefore, both tested methods aim to reduce the computational cost of optimizing such weights. Moreover, to further reduce the number of weights to find, the method performs a clustering in the feature space. This reduces the number of weights to find, as one weight is assigned per cluster. Another interesting finding reported by the authors, is the impact of \gls{OOD} data in batch normalization. Even if  the \gls{OOD} data lies far to the decision boundary, if batch normalization is carried out, a degradation of performance is likely. If no batch normalization is performed, \gls{OOD} data far from the decision boundary might not significantly harm performance. Therefore, the weights found are also used to perform a weighted mini-batch normalization of the data.  Regarding the benchmarking of the proposed method, the authors used the \gls{CIFAR-10} and Fashion\gls{MNIST} datasets, with different degrees of \gls{OOD} contamination. The \gls{OOD} data was sampled from a set of classes excluded from the \gls{IOD} dataset. A WRN-28-2 (WideResNet) backbone was used. No statistical analysis of the results, with the same number of partitions across the tested methods was performed. The average accuracy gains show a positive margin for the proposed method ranging from 5\% to 20\%.
 
 Another approach for robust \gls{SSDL} to distribution mismatch was proposed in \cite{wang2019semi}, referred to by the authors as \gls{ADA}. Similar to MixMatch, \gls{ADA} uses  MixMup \cite{zhang2017mixup} for data augmentation.  The method includes an unsupervised regularization term which measures the distribution divergence between the $S^{(u)}$ and $S^{(l)}$ datasets. This divergence is measured in the feature space. In order to diminish the empirical distribution mismatch of
 $S^{(u)}$ and $S^{(l)}$, the distribution distance of both datasets is minimized to build a  feature extractor
aiming for a  latent space where both feature densities are aligned. This is done through  adversarial loss optimization. We can categorize this method as a feature space based method.  As for the reported benchmarks, the authors did not test different degrees of \gls{OOD} data contamination, and only compared their method to generic \gls{SSDL} methods, not designed to handle distribution mismatch. No statistical significance tests were done to measure the confidence of their accuracy gains. In average, the proposed method seems to improve the error-rate from 0.5 to 2\%, when compared to other \gls{SSDL} generic methods. These results do not ensure a practical accuracy gain, as no statistical analysis was performed. From the baseline model with no distribution alignment, an accuracy gain of around 5\% was reported, again with no statistical meaning analysis performed. The authors in \cite{calderon2021dealing} also used the feature space to score unlabelled observations. The proposed method was tested in the specific application setting of COVID-19 detection using chest X-ray images. 

Recently, in \cite{huang2021trash}, an \gls{SSDL} approach to distribution mismatch robustness was developed. The method consists of two training steps. The first or \textit{warm-up } step  performs a self-training phase, where a pretext task is optimized using the \gls{DNN} backbone. This is implemented as the prediction of the degrees of rotation that each image, from a rotationally augmented dataset. This includes observations from both data samples $S^{(u)}$ and $S^{(l)}$, (along with the \gls{OOD} observations). In the second step, the model is trained using a consistency based regularization approach for the unlabelled data. This consistency regularization also uses the rotation consistency loss. In this step, an \gls{OOD} filtering method is implemented, referred by the authors as a cross-modal mechanism. This consists of the prediction of a pseudo-label, defined as the softmax of the \gls{DNN} output. This pseudo-label, along its feature embedding, is fed to what the authors refer to as a \textit{matching head}. Such \textit{matching head} consists of a multi-perceptron model that is trained to estimate whether the pseudo-label is accurately matched to its embedding. The matching head model is trained with the labelled data, with different matching combinations of the labels and the observations. As for the testing benchmark, the authors used \gls{CIFAR-10}, Animals-10 and CIFAR-ID-50 as \gls{IOD} datasets. For \gls{OOD}  data sources, images of Gaussian and Uniform noise, along with the \gls{TIN} and \gls{LSUN} datasets were used. The average accuracy reported for all the tested methods correspond to the last 20 model copies yielded during training. Therefore, no different training partitions were tested, preventing an  adequate statistical analysis of the results. The average results were not significantly better when compared to other generic \gls{SSDL} methods such as FixMatch, with an accuracy gain of around 0.5\% to 3\%. No computational cost figures about the cost of training the additional matching head or warm-up training were provided. 
The authors claim that in their method, \gls{OOD} data is re-used. However other methods like \gls{UASD} also prevent totally discarding \gls{OOD} data, as dynamic and soft observation-wise weights are calculated every epoch. Perhaps, from our point of view, a more appropriate description of the novelty of their method could be referred to as the complete usage of \gls{OOD} data in a pre-training stage.

\paragraph{Prior probability shift}\label{subsec:prior}
Data imbalance for supervised approaches  has been tackled in the literature widely.  Different approaches have been proposed, ranging from data transformations (over-sampling, data augmentation, etc.) to model architecture focused approaches (i. e., modification of thee loss function, etc.) \cite{ando2017deep,taherkhani2020adaboost,mullick2019generative}. Nevertheless, related to the problem of label imbalance or label balance mismatch between the labelled and unlabelled datasets, more scarce is found in the literature. This setting can be interpreted as a particularisation of the distribution mismatch problem described in \cite{oliver2018realistic}. A distribution mismatch between $S^{(l)}$ and $S^{(u)}$ might arise when the label or class membership distribution of the observations in both datasets meaningfully differ.

In \cite{hyun2020class}, an assessment of how distribution mismatch impacts a \gls{SSDL} model is carried out. The cause of the distribution mismatch between the labelled and unlabelled datasets was the label imbalance difference between them. An accuracy decrease between 2\% and 10\% was measured when the \gls{SSDL} faced such data setting.  The authors proposed a simple method to recover such performance degradation. The method consists on assigning a specific weight for each unlabelled observation in the loss term. To choose the weight, the output unit of the model with highest score at the current epoch is used as a label prediction. In this work, the mean teacher model was tested as a \gls{SSDL} approach \cite{tarvainen2017mean}. The authors yielded a superior performance of the \gls{SSDL} model by using the proposed method.   An extension to the work in \cite{hyun2020class} is found in \cite{calderon2021correcting}, where in this case the more recent MixMatch \gls{SSDL} method is modified to improve the robustness of the model to heavy imbalance conditions in the labelled dataset. The approach was extensively tested in the specific application of COVID-19 detection using chest X-ray images.

\section{Open challenges}\label{sec:conclusion}

Among the most important challenges faced by \gls{SSDL} under practical usage situations is the distribution mismatch between the labelled and unlabelled datasets. However, according to our state of the art review, there is significant work pending, mostly related to the implementation of standard benchmarks for novel methods. The benchmarks found so far in the literature show a significant bias towards the \textit{unseen class} distribution mismatch setting. No testing of other  distribution mismatch causes such as covariate shift was found in the literature. Real world usage settings might include covariate shift and prior probability distribution shift, which violate the frequently used \gls{IID} assumption. Therefore, we urge the community to focus on different distribution mismatch causes. 

Studying and developing methods for dealing with distribution mismatch settings, shifts the focus upon data-oriented (i.e., data transformation, filtering and augmentation) methods instead of more popular model-oriented methods. Recently, the renowned  researcher Andrew Ng, has drawn the attention towards data-oriented methods \footnote{\url{https://www.forbes.com/sites/gilpress/2021/06/16/andrew-ng-launches-a-campaign-for-data-centric-ai/?sh=68b63b2174f5}}. In his view,  not enough effort has been carried out by the community in studying and developing data-oriented methods to face  real-world usage settings. 

We agree with Andrew Ng's opinion, and add that besides establishing and testing a set of standard benchmarks where different distribution mismatch settings are tested, experimental reproducibility must be enforced. Recent technological advances not only allow to share the code and the datasets used, but also the testing environments through virtualization and container technology. Finally, we argue that the deep learning research community must be mindful of not only comparing average accuracies from the different state-of-the art methods. Statistical analysis tools must be used to test whether the performance difference between one method over another is reproducible and is statistically meaningful. Therefore we suggest that the results distribution is shared and not only the means and standard deviations of the results, in order to enable further statistical analysis.

\bibliographystyle{plain}
\bibliography{biblio}

\end{document}